\documentclass[lettersize,journal]{IEEEtran}
\usepackage{amsmath,amsfonts}
\usepackage{array}
\usepackage[caption=false,font=normalsize,labelfont=sf,textfont=sf]{subfig}
\usepackage{textcomp}
\usepackage{stfloats}
\usepackage{url}
\usepackage{verbatim}
\usepackage{graphicx}
\usepackage{cite}
\usepackage{enumerate}
\usepackage[backref]{hyperref}
\usepackage{multirow}
\usepackage{multicol}
\usepackage[table,xcdraw]{xcolor}
\usepackage{booktabs}
\usepackage{makecell}
\usepackage[flushleft]{threeparttable}
\usepackage[normalem]{ulem}
\useunder{\uline}{\ul}{}
 \usepackage{xcolor}
 \usepackage{multirow}
\usepackage[table,xcdraw]{xcolor}
\usepackage[flushleft]{threeparttable}
 \usepackage[ruled,linesnumbered]{algorithm2e}
 \definecolor{codered}{rgb}{0.98,0.49,0.72}
\begin{document}

\title{AMI-Net: Adaptive Mask Inpainting Network for Industrial Anomaly Detection and Localization}

\author{Wei Luo,~\IEEEmembership{Student Member,~IEEE,} Haiming Yao,~\IEEEmembership{Student Member,~IEEE,} \\Wenyong Yu,~\IEEEmembership{Senior Member,~IEEE}, and Zhengyong Li
        % <-this % stops a space
\thanks{This study was supported by the National Natural Science Foundation of China (Grant No. 52375494) (Corresponding author: Wenyong Yu.)}
\thanks{Wei Luo and Haiming Yao are with the State Key Laboratory of Precision Measurement Technology and Instruments, Department of Precision Instrument, Tsinghua University, Beijing 100084, China (e-mail: luow23@mails.tsinghua.edu.cn, yhm22@mails.tsinghua.edu.cn).}
\thanks{Wenyong Yu and Zhengyong Li are with the State Key Laboratory of Digital Manufacturing Equipment and Technology, School of Mechanical Science and Engineering, Huazhong University of Science and Technology, Wuhan 430074, China (e-mail: ywy@hust.edu.cn, m202370789@hust.edu.cn).}}

\markboth{IEEE Transactions on Automation Science and Engineering}%
{Shell \MakeLowercase{\textit{et al.}}: A Sample Article Using IEEEtran.cls for IEEE Journals}


\maketitle
\begin{abstract}
Unsupervised visual anomaly detection is crucial for enhancing industrial production quality and efficiency. Among unsupervised methods, reconstruction approaches are popular due to their simplicity and effectiveness. The key aspect of reconstruction methods lies in the restoration of anomalous regions, which current methods have not satisfactorily achieved. To tackle this issue, we introduce a novel \uline{A}daptive \uline{M}ask \uline{I}npainting \uline{Net}work (AMI-Net) from the perspective of adaptive mask-inpainting. In contrast to traditional reconstruction methods that treat non-semantic image pixels as targets, our method uses a pre-trained network to extract multi-scale semantic features as reconstruction targets. Given the multiscale nature of industrial defects, we incorporate a training strategy involving random positional and quantitative masking. Moreover, we propose an innovative adaptive mask generator capable of generating adaptive masks that effectively mask anomalous regions while preserving normal regions. In this manner, the model can leverage the visible normal global contextual information to restore the masked anomalous regions, thereby effectively suppressing the reconstruction of defects. Extensive experimental results on the MVTec AD and BTAD industrial datasets validate the effectiveness of the proposed method. Additionally, AMI-Net exhibits exceptional real-time performance, striking a favorable balance between detection accuracy and speed, rendering it highly suitable for industrial applications. Code is available at: \href{https://github.com/luow23/AMI-Net}{\textcolor{codered}{https://github.com/luow23/AMI-Net}}
\end{abstract}
\def\abstractname{Note to Practitioners}
\begin{abstract}
% Current defect detection methodologies exhibit proficiency in identifying defects present on textured surfaces. However, effectively identifying defects on objects with varying positional attributes remains a pronounced hurdle. This study presents AMI-Net as a solution for detecting and locating defects. It uses adaptive masking followed by inpainting techniques, eliminating the need for collecting extra defect samples for training. AMI-Net not only excels in identifying defects on textured surfaces but also extends its competency to encompass defects present in diverse items, including cable, transistor, capsule, and more. Furthermore, the detection speed of AMI-Net is exceptionally rapid, striking a harmonious equilibrium between detection precision and speed, which is a facet bearing immense significance in the realm of authentic industrial defect detection.
AMI-Net restores defective images to normal ones and subsequently detects defects by leveraging the differences between them. This method only needs to collect about a few hundred defect-free samples for training, without the need for additional defect samples. It is noteworthy that AMI-Net is applicable not only to the detection of simple texture surface defects, such as carpet, leather, and tile, but also to the detection of surface defects in objects with posture diversity, such as cable, transistor, and screw. The trained model not only exhibits high detection accuracy but also demonstrates superior real-time performance, showcasing significant potential in practical industrial settings.
\end{abstract}

\begin{figure}
    \centering
    \includegraphics[width=88mm]{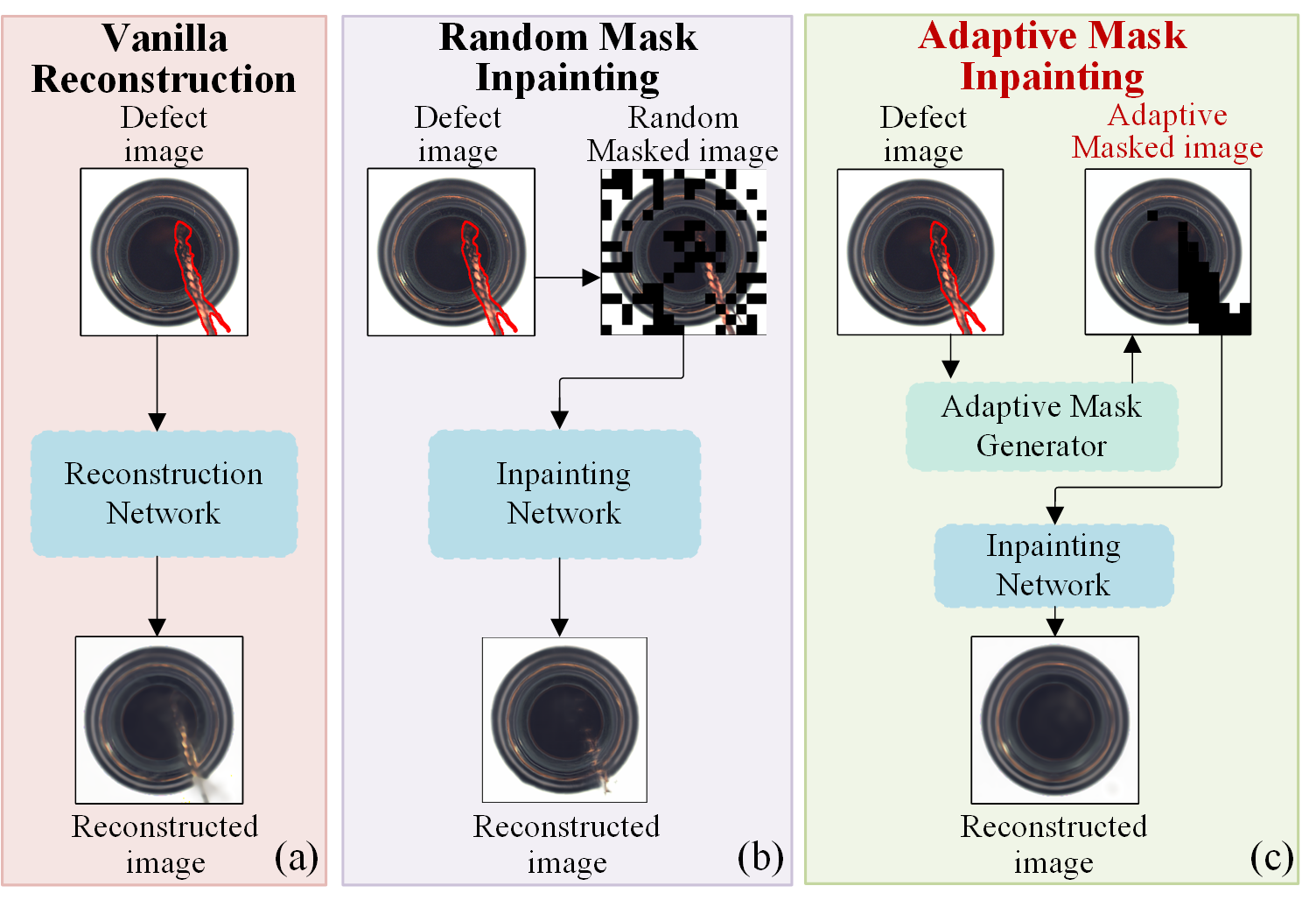}
    \caption{Comparison of different unsupervised anomaly detection methods. (a) Vanilla autoencoder \cite{AE}. (b) Existing mask-based method \cite{RIAD}. (c) The proposed method (AMI-Net). It is noteworthy that our model is learned for feature reconstruction and a separate decoder is employed to render images from features. This decoder is only used for visualization.}
    \label{fig:intro}
\end{figure}
\section{Introduction}
\IEEEPARstart{I}{n} industrial manufacturing processes, a myriad of uncontrollable factors can give rise to defects in industrial products, thereby compromising their intrinsic value. Consequently, industrial anomaly detection \cite{MCDEN, DCD, pmemory} assumes a pivotal role in the realm of industrial quality supervision.\\
\indent Owing to the unpredictable nature of industrial defects and the significantly higher number of normal samples in comparison to abnormal samples, the collection and annotation of a comprehensive dataset are deemed impractical. This greatly restricts the application of traditional supervised learning \cite{PGANet, A-Net} methods. As a result, a substantial amount of research \cite{NDP-Net, FMR-Net, CMA-AE} on anomaly detection is conducted under the unsupervised paradigm.\\
\indent Unsupervised anomaly detection methods are built upon the premise of utilizing only normal samples for training. During the inference process, regions that deviate significantly from the normal patterns are identified as anomalies. The reconstruction-based methods have garnered widespread attention due to their simplicity and effectiveness. As a typical example of reconstruction methods, autoencoder (AE) \cite{AE} establishs a representation of the distribution of normal samples by minimizing the reconstruction error, which refers to the disparity between input images and corresponding reconstructed images. 
AE posits that a model trained only on normal data is capable of reconstructing normal patterns during testing but struggles to accurately reconstruct abnormal patterns, resulting in larger reconstruction errors for abnormal samples and effectively distinguishing them from normal samples. However, this assumption does not always hold true, as neural networks exhibit strong generalization capabilities. Specifically, as shown in Fig. \ref{fig:intro}(a), AE can accurately reconstruct not only normal regions but also defects during the testing process. This results in a minimal difference in reconstruction errors between normal and abnormal regions, making it challenging to differentiate between abnormal and normal samples.\\
\indent In order to address the issue in AE that defects are also well reconstructed, several mask-based approaches \cite{RIAD, self-mask,jiang2022masked} have been proposed, transforming the image reconstruction task into an image inpainting task. However, existing mask-based methods exhibit certain limitations. Firstly, they employ random masks during testing, which may inadequately mask all defect regions. This results in the model leveraging semantic information from unmasked defect areas to generate defect patterns in the masked regions, as illustrated in Fig. \ref{fig:intro}(b).  Additionally, employing random masks during testing introduces instability in model performance. In other words, varied random masks result in disparate detection outcomes, posing a critical challenge for real-world industrial scenarios that demand exceptionally high stability. Moreover, these methods adopt multiple complementary masks during the testing phase, causing a notable decrease in the model's inference speed. In industrial scenarios where strict real-time requirements exist, this limitation hinders the practical viability of these methods for actual industrial applications.\\
\indent To tackle the aforementioned limitations in existing mask-based methods, we introduce a novel adaptive mask inpainting network (AMI-Net) for unsupervised industrial anomaly detection. Diverging from conventional image reconstruction approaches, we utilize features extracted by a pre-trained network as the reconstruction targets. During the training phase, we still employ a random position and quantitative masking strategy. However, unlike conventional mask-based methods, we introduce a variable masking ratio instead of a fixed one, which refers to the proportion or percentage of the masked
area in relation to the total area of the image. This adjustment is motivated by the diverse scales inherent in real industrial defects. Subsequently, during the testing phase, we propose a novel adaptive mask generator, which dynamically generates adaptive masks for each image, masking the defective regions while retaining the normal regions. In this manner, as shown Fig. \ref{fig:intro}(c), the model can leverage the visible normal global contextual information to restore the masked defective regions, thereby effectively suppressing the reconstruction of defects. Furthermore, our model avoids randomness during the testing process and eliminates the need for multiple complementary masks, thereby enhancing the stability and real-time performance of the model. This characteristic demonstrates significant potential in real-world industrial scenarios. The main contributions of this article can be summarized as follows:
\begin{itemize}
    \item We propose a novel mask-based reconstruction method, AMI-Net. This approach employs a random positional and quantitative masking strategy to address the multi-scale issue of defects in industrial scenarios. In the testing phase, we introduce a novel adaptive mask generator capable of adaptively masking defect areas while preserving normal regions as much as possible. This effectively addresses the issue of perfect defect reconstruction in the testing process for existing mask-based methods.
    \item The proposed AMI-Net demonstrates outstanding applicability in the industrial domain. Firstly, it exhibits remarkable inspection performance on two industrial datasets, MVTec AD \cite{MVTEC} and BTAD \cite{BTAD}. Furthermore, under three different training settings, one-for-one (training separate models for different categories of objects), one-for-all (training a unified model for detecting anomalies across all different object categories), and few-shot (training models using a few samples), AMI-Net consistently achieves excellent detection performance. Finally, the inference time of AMI-Net is measured at 11.48 ms, highlighting its superior real-time performance. Therefore, AMI-Net holds significant potential for widespread application in industrial scenarios.
\end{itemize}

The remaining structure of this paper is organized as follows: A review of related research on unsupervised industrial anomaly detection is presented in Section \MakeUppercase{\romannumeral2}. Section \MakeUppercase{\romannumeral3} provides a detailed introduction to AMI-Net. Extensive comparative and ablation analyses are conducted in Section \MakeUppercase{\romannumeral4}. Finally, Section \MakeUppercase{\romannumeral5} concludes the paper with a summary and a discussion of future research directions.
% \begin{figure*}[!t]
%     \centering
%     \includegraphics{pic/03.png}
%     \caption{Caption}
%     \label{fig:enter-label}
% \end{figure*}

\section{Related Work}
\subsection{Reconstruction-based method}
In this subsection,  we present an overview of reconstruction-based methods from the perspective of data usage. Depending on the type of data employed, we categorize these methods into three groups: those based on normal data, those based on artificial anomalies, and mask-based methods. 
\subsubsection{Reconstruction method using normal data}
\begin{figure}[!t]
    \centering
    \includegraphics{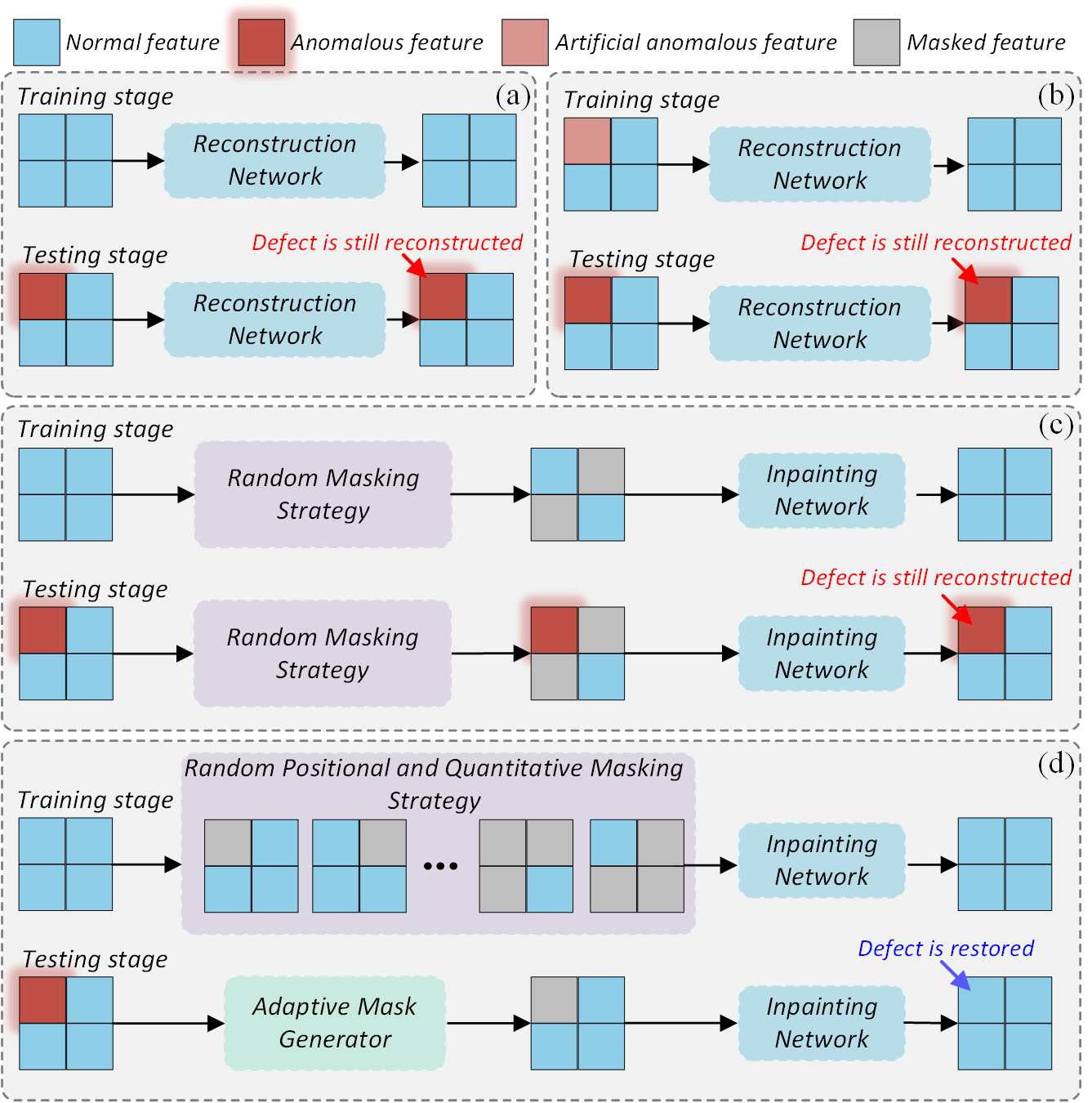}
    \caption{The schematic diagrams of different methods. (a) Normal-data-based method utilizes only normal samples for training. However, during the testing phase, it still reconstructs defects, as neural networks inherently possess the property of generalization. (b) Artificial-defect-based method employs artificial defect samples for training. However, due to the lack of authenticity in artificial defects, real defects continue to be reconstructed during the testing phase. (c) Existing mask-based method employs a random masking strategy during the training process. However, during the testing phase, random masks fail to completely mask the defective areas, resulting in the reconstruction of defects. (d) Our method employs a random positional and quantitative masking strategy during the training process. During the testing phase, it generates the adaptive masks for defect images, effectively concealing all defect regions and achieving defect restoration.}
    \label{fig:02}
    \vspace{-2em}
\end{figure}

As depicted in Fig. \ref{fig:02}(a), reconstruction methods based on normal data refer to approaches that exclusively use normal samples for the reconstruction task, without incorporating other data augmentation techniques, such as artificial anomalies or mask-based strategies. Autoencoders (AE) \cite{AE} minimize the reconstruction error of normal samples during the training process, and larger reconstruction errors are expected for the regions with anomalies during the testing process. However, the powerful generalization capability of neural networks leads to AE reconstructing defects, resulting in relatively smaller reconstruction errors for the anomalous regions. To address this issue, numerous methods \cite{MemAE, TrustMAE, DAAD, MSCDAE, MSFCAE} have been proposed. MemAE \cite{MemAE} introduces a memory module that utilizes normal memory items to replace anomalous features, thereby alleviating the problem of defect reconstruction. TrsutMAE \cite{TrustMAE} improves the memory mechanism and leverages perceptual distance \cite{perceptual_distance} to jointly enhance the detection accuracy. DAAD \cite{DAAD} proposes a block-wise memory module from the perspective of feature separation and assembly. Indeed, while these methods have effectively improved performance, the approaches based on normal data still tend to reconstruct defect regions.
\subsubsection{Reconstruction method using artificial defect}
As shown in Fig. \ref{fig:02}(b), reconstruction methods based on artificial anomalies involve designing artificial defects to simulate real defects in industrial scenarios, thereby addressing the issue of defect reconstruction. Cutpaste \cite{li2021cutpaste} involves randomly cutting a portion of the original image and then randomly pasting it back onto the image. AFEAN \cite{AFEAN} designs artificial defects by introducing random grayscale variations in specified regions. Considering the diverse shapes of defects, DRAEM \cite{draem} incorporates a Perlin noise to define the shapes of artificial anomalies. Furthermore, natural images contain redundant features relative to industrial images. Therefore, DRAEM employs natural image blocks in designated regions to emulate real industrial defects, resulting in a notable improvement in the model's performance. However, despite the many advancements in methods \cite{MLDFR, NDP-Net, FMR-Net} based on artificial anomalies, it remains impractical to simulate all real-world defects using artificial anomalies due to the unpredictable nature of defects. Furthermore, relying solely on artificial anomalies can lead to model overfitting and a lack of generalization.
\subsubsection{Reconstruction method using mask strategy}
As shown in Fig. \ref{fig:02}(c), reconstruction methods based on mask strategy approach anomaly detection from an image inpainting perspective. They utilize information from normal regions to repair the masked regions. RIAD \cite{RIAD} employs multi-scale random masking to enhance the model's ability to perform inpainting. On the other hand, ST-MAE \cite{STMAE} utilizes complementary random masks for feature transformation, which leads to meaningful reconstructed outcomes. MSTUnet \cite{jiang2022masked} utilizes anomaly generation and masking techniques, incorporating the Swin Transformer \cite{swintransformer} as the inpainting network, resulting in remarkable performance. However, the aforementioned mask-based methods use random masking during the testing phase, which has the potential to leave defect areas inadequately covered and thus, may result in false positives. SSM \cite{self-mask} employs a progressively refined incremental masking approach, progressively identifying normal regions and ultimately identifying anomalous areas. InTra \cite{Intra} reconstructs the image by iteratively sliding a mask and eventually concatenating all repaired masked patches. While these two methods do not involve random masking, their reliance on multiple iterations to obtain detection results leads to a significant decrease in inference speed. Therefore, they are not suitable for industrial applications. As illustrated in Fig. \ref{fig:02}(d), our approach introduces an adaptive mask generator to alleviate the issue of defects still being reconstructed. Furthermore, our method does not require multiple iterations; it runs only once to obtain detection results, significantly enhancing inference speed and demonstrating substantial potential in industrial scenarios.

\subsection{Feature embedding-based method}
Feature embedding-methods construct a feature embedding space through the utilization of a substantial volume of normal data, wherein the embedding distance of anomalous features surpasses that of normal features. Deep SVDD \cite{DeepSVDD} utilizes a neural network to map normal samples onto the smallest volume hypersphere, thus categorizing samples outside this hypersphere as anomalies. On the other hand, Patch SVDD \cite{PatchSVDD} focuses on the patch level, achieving anomaly localization. SPADE \cite{SPADE} and PatchCore \cite{PacthCore} conducts anomaly detection by computing distances between pre-stored typical normal features and test features. GCPF \cite{GCPF} utilizes a multivariate Gaussian distribution for modeling and employs the Mahalanobis distance as the anomaly criterion. 
RD4AD \cite{RD4AD} attains exceptional anomaly detection performance by employing reverse knowledge distillation. MBPFM \cite{MBPFM} leverages two distinct pre-trained networks for feature extraction and enables mutual feature mapping, resulting in robust performance. While the aforementioned feature embedding methods achieve impressive results, they often come with significant computational overhead, resulting in notably slow inference speed.

\section{The AMI-Net Methodology}
In this section, we provide a comprehensive explanation of the proposed AMI-Net. We begin by presenting the overarching framework of AMI-Net. Subsequently, we delve into the details of the multi-scale pre-trained feature extraction. Moving forward, we introduce the strategy of employing a random positional and quantitative masking approach. Following this, the adaptive mask generator is introduced. Furthermore, we provide insights into both the training and inference procedures of the network. Lastly, we outline the specific parameter configuration utilized in the network.
\begin{figure*}
    \centering
    \includegraphics{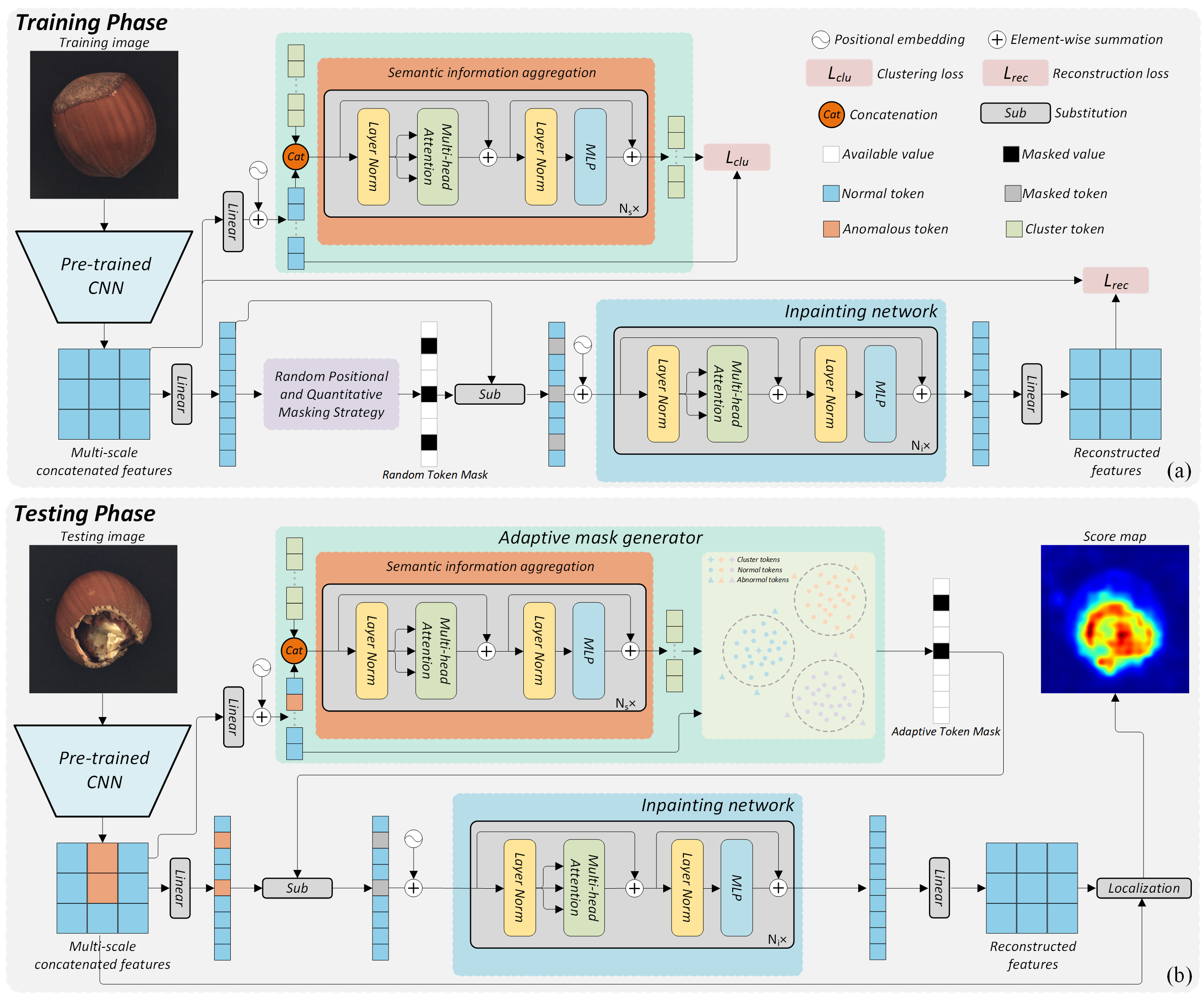}
    \caption{Overall architecture of proposed AMI-Net. Firstly, multi-scale features are extracted using a pretrained CNN. (a) During the training phase, AMI-Net employs a strategy involving randomized positions and quantities of masks for the inpainting task. (b) During the testing phase, AMI-Net employs an adaptive mask generator to create a mask that dynamically conceals the defective region while preserving the normal area. Subsequently, the inpainting network is applied to acquire a reconstructed feature that is devoid of anomalies. Ultimately, by analyzing the input alongside the reconstructed feature, defects can be accurately localized.}
    \label{fig:03}
    % \vspace{-1em}
\end{figure*}

\subsection{Overall Framework of AMI-Net}
The overall architecture of AMI-Net is illustrated in Fig. \ref{fig:03}. Initially, multi-scale features are extracted using a pre-trained convolutional neural network (CNN), with these features serving as the reconstruction targets instead of the complete images. During the training phase, a random positional and quantitative masking strategy are employed, allowing the model to acquire the capability to handle various types of defects. Additionally, in the training process, cluster tokens are introduced to aggregate normal features, thereby laying the foundation for the adaptive masking process during inference.\\
\indent In the testing phase, an adaptive mask generator is employed to generate adaptive masks for the multi-scale features. These masks obscure defect-related features while retaining normal features. Subsequently, an inpainting network is utilized to generate defect-free reconstructed features. Finally, the defect localization map is derived by utilizing both the input features and the reconstructed features.\\
\indent The comprehensive architecture of AMI-Net synergistically leverages these components to proficiently enable the tasks of anomaly detection and localization. The inclusion of adaptive masks and the integration of cluster tokens synergistically enhance the model's adaptability in handling a wide range of anomalies, thereby significantly bolstering its overall performance.
\vspace{-2pt}
\subsection{Multi-Scale Pre-Trained Feature Extraction}
Reference \cite{MKD} has provided evidence that normal and abnormal patterns exhibit heightened discriminative characteristics within the feature space. Therefore, we leverage the WideResnet50 \cite{wideresnet} $\phi$ pre-trained on ImageNet \cite{ImageNet} to extract features. WideResNet50 proves adept at extracting abundant semantic information while maintaining a relatively fast inference speed, striking a favorable balance between detection performance and inference speed, as validated by the ablation experiments in Section \ref{pre-trianednetwork}. Given an image $I\in R^{H\times W\times C}$, we can obtain feature maps at different scales $\{\phi_{1}(I), \phi_{2}(I), \phi_{3}(I), \phi_{4}(I)\}$. Feature maps at different scales carry distinct semantic information, with richer semantic content found in deeper layers. Considering the diverse scales of defects, we scale the feature maps from the $2_{nd}$ to the $4_{th}$ layers to the same size and then concatenate them along the channel dimension, yielding a multi-scale concatenated feature representation $\mathcal{F}(I)\in R^{H_F\times W_F\times C_F}$.
\begin{equation}
    \mathcal{F}(I) = \mho \{\Theta(\phi_{2}(I)), \Theta(\phi_{3}(I)), \Theta(\phi_{4}(I))\}
\end{equation}
Where $\Theta$ signifies the resizing of feature maps to dimensions of $H_F\times W_F$ and $\mho$ represents the operation of concatenating features along the channel dimension.

\subsection{Random Positional and Quantitative Masking Strategy}
\label{Lrec}
As illustrated in Fig. \ref{fig:03}(a), following the acquisition of multi-scale features, we employ a linear projection $f_{linear}^1$ with a patch size of $K$ to transform the 2D feature map $\mathcal{F}(I)$ into a 1D sequence of feature tokens $E_{\mathcal{F}(I)}^1=\{T_{\mathcal{F}(I)}^{1\-1},T_{\mathcal{F}(I)}^{1\-2},\cdots,T_{\mathcal{F}(I)}^{1\-L}|T_{\mathcal{F}(I)}^{1\-i}\in R^D\}$.
\begin{equation}
    E_{\mathcal{F}(I)}^1=f_{linear}^1(\mathcal{F}(I); \theta_{linear}^1)
\end{equation}
 In the above context, $E_{\mathcal{F}(I)}^1\in \mathbb{R}^{L\times D}$, where $L=\frac{H_F}{K}\times \frac{W_F}{K}$, and $f_{linear}^1$ as well as $\theta_{linear}^1$ correspondingly denote the function and parameters of the linear projection. The $D$ signifies the feature dimension of the tokens.\\
\indent MAE \cite{he2022masked} uses a fixed masking ratio to train the visual representation ability of the model. However, employing a fixed masking ratio for anomaly detection is unsuitable, as industrial defects can vary greatly in size. To address this limitation, our study introduces a dynamic masking ratio, randomly applied, to effectively handle defects of diverse sizes. Initially, we randomly sample a mask ratio from a uniform distribution ranging between 0 and 1. Subsequently, based on the chosen mask ratio, we generate the corresponding mask $M_{random}\in R^L$ for subsequent procedures. Provided with a feature token $T_{mask}\in R^D$ comprising exclusively zeros, our next step involves masking the specified region:
\begin{equation}
    \Tilde{E}_{\mathcal{F}(I)}^1=Sub(E_{\mathcal{F}(I)}^1, M_{random}, T_{mask})
\end{equation}
Where $Sub$ operation denotes replacing the tokens in $E_{\mathcal{F}(I)}^1$ at positions where the corresponding positions in matrix $M_{random}$ are 0 with the $T_{mask}$. Then, we add fixed sinusoidal positional embedding $E_{pos}$ into $\Tilde{E}_{\mathcal{F}(I)}^1$:
\begin{equation}
    \Tilde{E}_{\mathcal{F}(I)}^1 = \Tilde{E}_{\mathcal{F}(I)}^1 + E_{pos}
\end{equation}
Where $E_{pos}\in R^{L\times D}$. Subsequently, $\Tilde{E}_{\mathcal{F}(I)}^1$ is fed into the inpainting network $f_{in}$ to obtain the reconstructed feature token sequence $\hat{E}_{\mathcal{F}(I)}^1$.
\begin{equation}
    \hat{E}_{\mathcal{F}(I)}^1 = f_{in}(\Tilde{E}_{\mathcal{F}(I)}^1;\theta_{in})
\end{equation}
Where $f_{in}$ and $\theta_{in}$ respectively represent the function and parameters of the inpainting network. As shown in Fig. \ref{fig:03}(a), the inpainting network consists of $N_{i}$ transformer blocks, where each transformer block includes layer normalization (LN), multi-head self-attention mechanism (MSA), and multi-layer perceptron (MLP). Subsequently, we utilize an additional linear projection $f_{linear}^2$ to convert the 1D token sequence $\hat{E}_{\mathcal{F}(I)}^1$ into a 2D feature map $\hat{F}(I)$.
\begin{equation}
    \hat{\mathcal{F}}(I)=f_{linear}^2(\hat{E}_{\mathcal{F}(I)}^1;\theta_{linear}^{2})
\end{equation}
Where $f_{linear}^2$ and $\theta_{linear}^2$ represent the function and parameters of the linear projection respectively.\\
\indent  In image reconstruction tasks, the mean squared error (MSE) is frequently employed as a reconstruction loss metric. In addition to MSE, cosine similarity is also utilized to gauge feature similarity. As a result, within our research, we opt for a combination of both cosine similarity and MSE as reconstruction loss functions.
\begin{equation}
    L_{mse} = \mathbb{E}_{P_{I}}\left[ {\big\|\hat{\mathcal{F}}(I)-\mathcal{F}(I)\big\|}^2\right]
\end{equation}
\begin{equation}
     L_{cos} = \mathbb{E}_{P_{I}}\left[1-\frac{\hat{\mathcal{F}}(I)\cdot\mathcal{F}(I)}{\big\|\hat{\mathcal{F}}(I)\big\|\times \big\|\mathcal{F}(I)\big\|}\right]
\end{equation}
\begin{equation}
    L_{rec} = w_1L_{mse} + w_2L_{cos}
\end{equation}
Where ${\big\|\cdot\big\|}^2$, ${\big\|\cdot\big\|}$, $w_1$ and $w_2$ denote the mean squared error, the modulus, and the weights assigned to $L_{mse}$ and $L_{cos}$, respectively. Cosine similarity is more accurate for comparing feature similarity than MSE. Therefore, in this study, $w_1$ and $w_2$ are set to 1 and 5, respectively.

\begin{figure*}[!t]
    \centering
    \includegraphics[width=183mm]{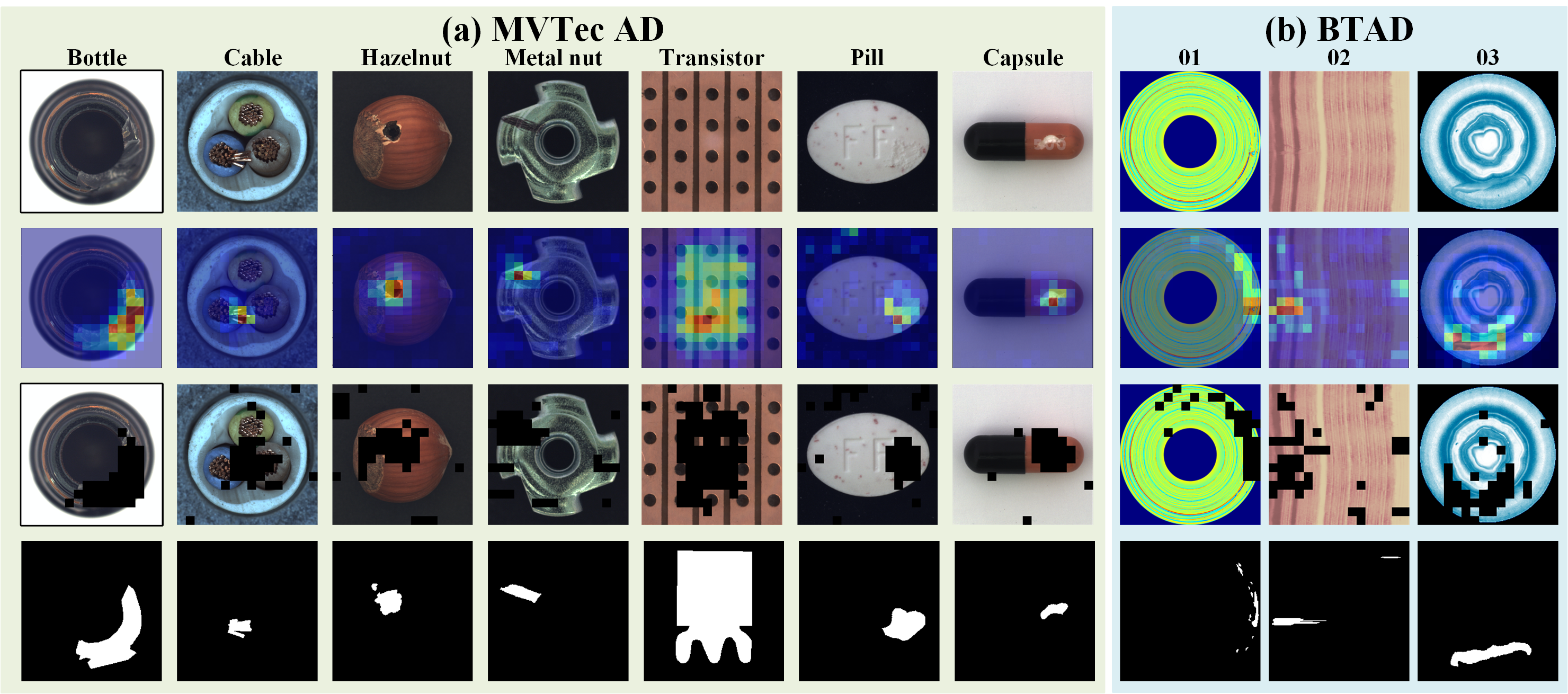}
    \caption{Examples of the effectiveness of the adaptive mask generator. First Row: the defective image. Second Row: the distance map formed by the distance between latent feature and their corresponding cluster centers. Third Row: the adaptive mask. Final Row: the corresponding label.}
    \label{fig:04}
\end{figure*}
\subsection{Adaptive Mask Generator}
\label{Lclu}
As depicted in Fig. \ref{fig:02}(c), existing mask-based methodologies also involve the application of random masks during the testing phase, thus leading to situations where defective regions are not effectively concealed. In order to tackle this issue, we introduce a novel adaptive mask generator. Moving forward, we provide a comprehensive overview of the adaptive mask generator for both training and testing perspectives, delving into specific details
\subsubsection{Training Stage}
As illustrated in Fig. \ref{fig:03}(a), we utilize an additional linear projection $f_{linear}^3$ to convert the multi-scale features $\mathcal{F}(I)$ into a 1D token sequence $E_{\mathcal{F}(I)}^2=\{T_{\mathcal{F}(I)}^{2\-1},T_{\mathcal{F}(I)}^{2\-2},\cdots,T_{\mathcal{F}(I)}^{2\-L}|T_{\mathcal{F}(I)}^{2\-i}\in R^D\}$:
\begin{equation}
     E_{\mathcal{F}(I)}^2=f_{linear}^3(\mathcal{F}(I); \theta_{linear}^3)
\end{equation}
Where $f_{linear}^3$ and $\theta_{linear}^3$ respectively represent the function and parameters of the linear projection, and the sequences of $E_{\mathcal{F}(I)}^2$ and $E_{\mathcal{F}(I)}^1$ share the same length and feature dimension. Then, we add fixed positional embedding $E_{pos}$ into $E_{\mathcal{F}(I)}^2$:
\begin{equation}
    E_{\mathcal{F}(I)}^2=E_{\mathcal{F}(I)}^2+E_{pos}
\end{equation}
\indent Subsequently, we introduce $P$ trainable cluster tokens to form a sequence of cluster tokens $E_{clu}=\{T_{clu}^1,T_{clu}^2,\cdots,T_{clu}^P|T_{clu}^i\in R^D\}$. Then, we concatenate $E_{clu}$ and $E_{\mathcal{F}(I)}^2$ along the dimension of length and feed them into semantic aggregation network $f_{sem}$, enabling the semantic information from $E_{\mathcal{F}(I)}^2$ to be aggregated into $E_{clu}$. The semantic aggregation network is composed of $N_s$ transformer blocks. Subsequently, we employ $E_{clu}$ and $E_{\mathcal{F}(I)}^2$ to define the clustering loss $L_{clu}$. As depicted in Fig. \ref{fig:03}(a), it is worth noting that the $E_{\mathcal{F}(I)}^2$ utilized for clustering is the one with added positional embedding but not influenced by $f_{sem}$. Currently, two widely utilized distance metrics are the Euclidean distance and cosine similarity. The former only takes into account the absolute distance between feature vectors without considering their direction, while the latter focuses solely on direction and neglects magnitude. Therefore, to achieve a more comprehensive measurement of the distance between feature vectors, we integrate these two metrics to attain a more holistic evaluation. The distance between $T_{clu}^i$ and $T_{\mathcal{F}(I)}^{2\-j}$ is defined as follows:
\begin{equation}
    R(T_{clu}^i,T_{\mathcal{F}(I)}^{2j}) = \underbrace{{\big\|T_{clu}^i-T_{\mathcal{F}(I)}^{2j}\big\|}^2}_{Euclidean\;distance}\times \underbrace{(1-\frac{T_{clu}^i\cdot T_{\mathcal{F}(I)}^{2j}}{\big\|T_{clu}^i\big\|\times \big\|T_{\mathcal{F}(I)}^{2j}\big\|})}_{{Cosine\;similarity}}
\end{equation}
Where $i=1,\cdots,P$, $j=1,\cdots, L$, and a smaller value of $R$ indicates a higher similarity between $T_{clu}^i$ and $T_{\mathcal{F}(I)}^{2\-j}$. For each cluster token, the latent feature $T_{\mathcal{F}(I)}^{2\-j}$ has a corresponding distance. We assign $T_{\mathcal{F}(I)}^{2\-j}$ to the cluster token with the smallest distance.
\begin{equation}
    d_{ij}=\min_{i\in \{1,\cdots,P\}}R(T_{clu}^i,T_{\mathcal{F}(I)}^{2j})
\end{equation}
\begin{equation}
    d_{i}=\sum_{j}d_{ij}
\end{equation}
Where $d_{ij}$ is a set that includes latent feature tokens belonging to the $i_{th}$ cluster, and $d_i$ is the sum of distances from all latent feature tokens belonging to the $i_{th}$ cluster to the cluster center. The objective of clustering is to minimize intra-class variance while maximizing inter-class variance. Reducing intra-class variance leads to more concentrated and compact features, while increasing inter-class variance ensures feature diversity. Therefore, the clustering loss $L_{clu}$ is defined as follows:
\begin{equation}
    L_{clu} = \mathbb{E}_{P_I} \left[\underbrace{w_3\sum_{i}^Pd_i}_{intra-class}-\underbrace{w_4\sum_{i}^{P}\sum_{j}^{P}R(T_{clu}^i, T_{clu}^j)}_{inter-class} \right]
\end{equation}
Where $w_3$ and $w_4$ are the weights assigned to the first and second terms, set to 1 and 0.1 respectively.
\begin{figure}[!t]
    \centering
    \includegraphics[width=85mm]{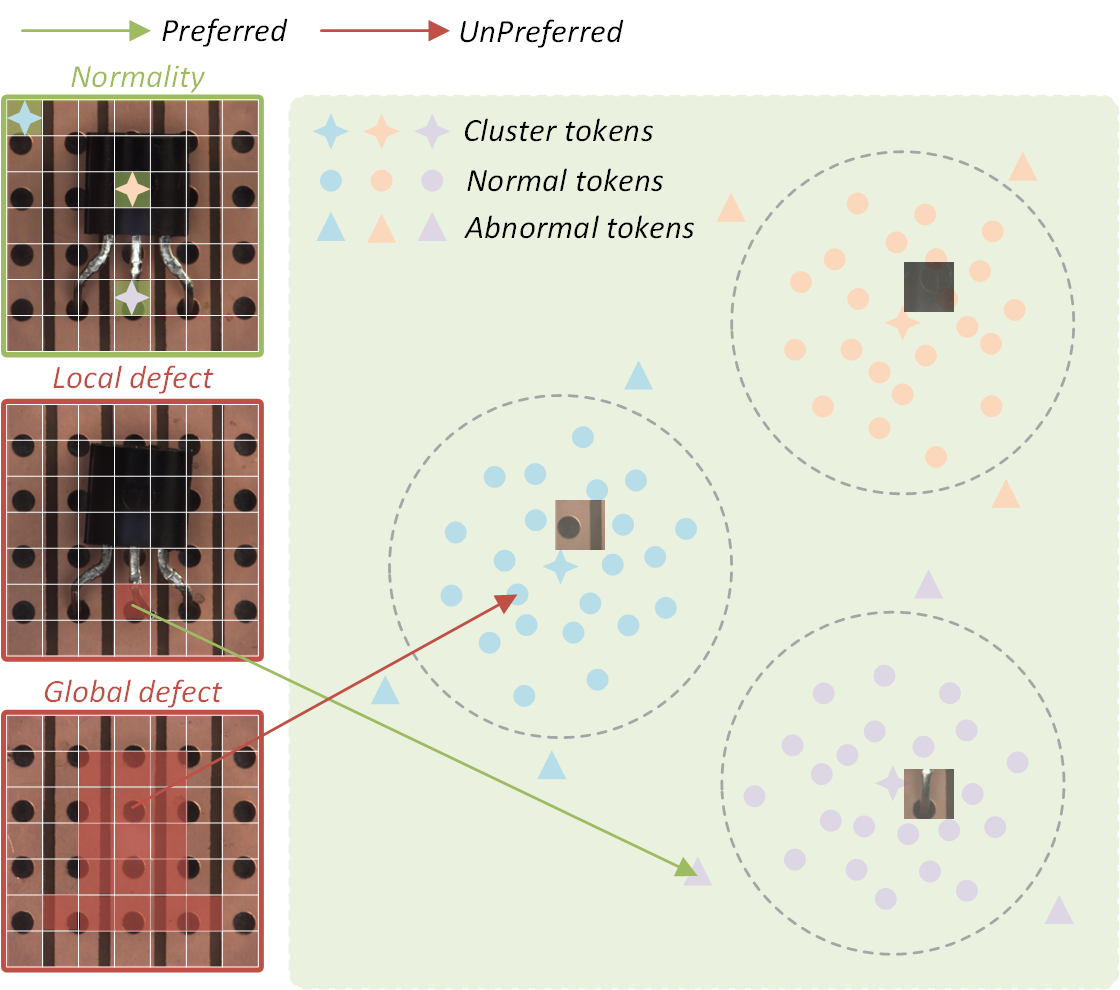}
    \caption{Issues arising from clustering methods that do not take positional information into account.}
    \label{fig:05}
    \vspace{-1.0em}
\end{figure}
\subsubsection{Testing Stage}
The trained adaptive mask generator can generate adaptive masks during the testing phase for different token sequences. These masks effectively conceal defective regions while preserving normal areas.\\
\indent As discussed above, upon obtaining the distance set $d_{ij}$ for the $i_{th}$ cluster, we can proceed to define the boundary of the $i_{th}$ cluster.
\begin{equation}
     r_{i}=\varpi\left(d_{i j}\right)+\lambda \sigma\left(d_{i j}\right)
\end{equation}
Where $\varpi$ and $\sigma$ denote the operations of calculating mean and standard deviation, and $\lambda$ is the scaling factor used to control the cluster boundary. In our study, the value of $\lambda$ is set to 0.5.\\
\indent Following this, within the $i_{th}$ cluster, latent feature tokens that surpass the boundary $r_i$ are masked, while those within the boundary $r_i$ are preserved. By applying the same procedure to all clusters, the adaptive mask $M_{adaptive}$ can be generated. Subsequently, the inpainting network utilizes the available normal information to restore the masked regions, resulting in a defect-free reconstructed feature.\\
\indent Fig. \ref{fig:04} illustrates the effectiveness of the adaptive mask generator. Our introduced adaptive mask generator proficiently conceals defect regions while preserving normal areas.
\subsubsection{Reason for incorporating an additional linear projection $f_{linear}^3$}
In the event that $f_{linear}^3$ and $f_{linear}^1$ share weights, the optimization of both $L_{rec}$ and $L_{clu}$ would impact the $f_{linear}^1$, resulting in a mutual influence between the adaptive mask generator and the inpainting network. This, in turn, could lead to training instability. To address this concern, we introduce an additional linear projection $f_{linear}^3$, wherein $L_{rec}$ focuses on optimizing $f_{linear}^1$ and $L_{clu}$ concentrates on optimizing $f_{linear}^3$. By adopting this approach, we effectively enhance the model optimization process and alleviate potential instability problems.
\begin{figure}[!t]
    \centering
    \includegraphics[width=85mm]{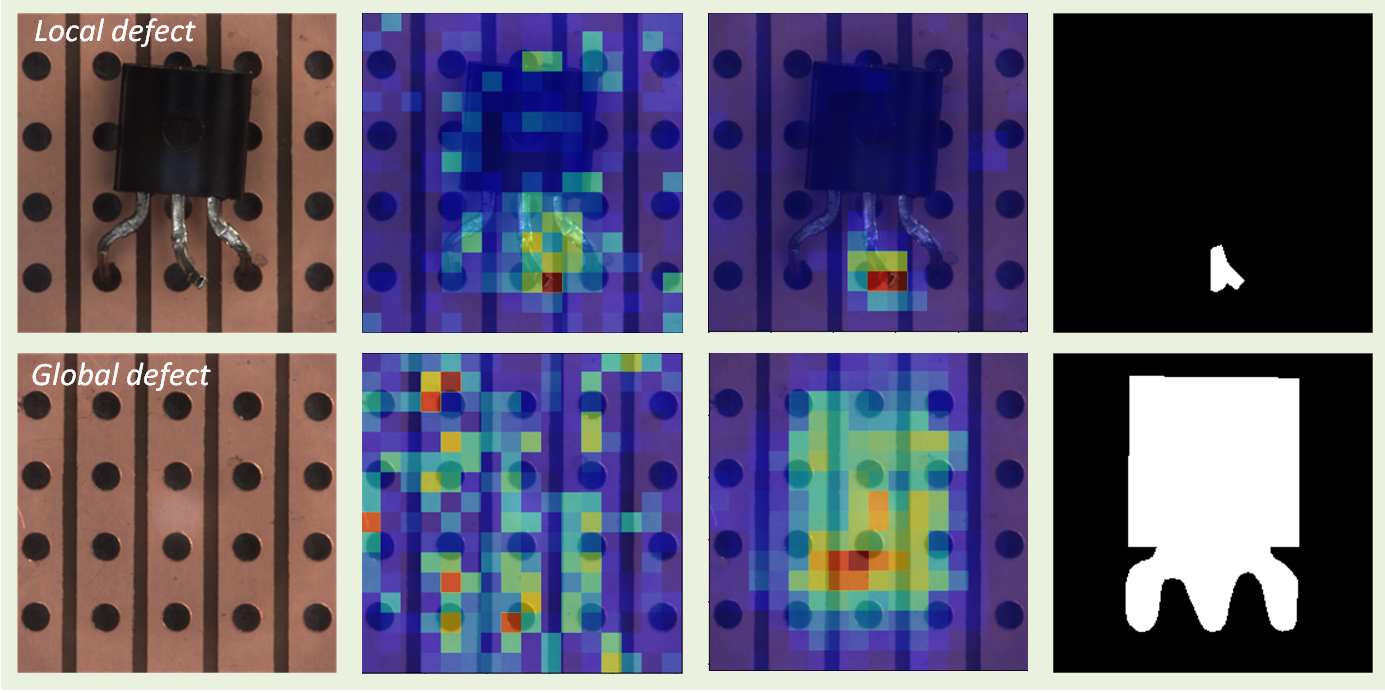}
    \caption{Examples of the role of positional information. First Column: the defective image. Second Column: the distance map without positional information. Third Column: the distance map with positional information. Final Column: the corresponding label.}
    \label{fig:06}
\end{figure}
\subsubsection{Reason for employing tokens with added positional embedding for clustering}
Some methods \cite{AFEAN,ACDN,MSFCAE} directly employ latent features for clustering, but this can give rise to certain issues. For instance, as depicted in Fig. \ref{fig:05}, considering the Transistor dataset within the MVTec AD dataset \cite{MVTEC}, let's assume three clusters exist. For local defects, they can be easily discerned since local defects are generally structural damages, rendering them distinctly dissimilar from normal features. However, for global defects, they are likely classified as part of the cluster boundaries, owing to the fact that global defects often involve object disappearance and disorder of order. Consequently, global defects appear similar to normal patterns in terms of features, but their positions differ. Hence, the inclusion of positional information for clustering is deemed essential. Fig. \ref{fig:06} illustrates the significance of incorporating positional information for clustering.
\subsubsection{Reason for selecting ViT as the semantic information aggregation network and inpainting network}
The semantic information aggregation task aims to aggregate information from input images onto cluster tokens, while the inpainting task aims to utilize visible contextual information to restore masked regions. Both tasks crucially require semantic information with a global context. ViT, owing to its self-attention mechanism that considers correlations between every image patches, exhibits global modeling capabilities. In contrast, CNNs encompass two inductive biases, locality and translational equivariance, restricting them to consider semantic information only within fixed convolutional kernel regions, lacking global modeling capabilities. Therefore, we opted for ViT as the semantic information aggregation network and inpainting network.
\subsection{Training and Inference Processes}
AMI-Net comprises two core components: the adaptive mask generator and the inpainting network. The clustering loss $L_{clu}$ (in Section \ref{Lclu}) is employed for optimizing the former, while the reconstruction loss $L_{rec}$ (in Section \ref{Lrec}) is applied to enhance the latter. Hence, the holistic loss function of AMI-Net is defined as follows:
\begin{equation}
    L_{total} = L_{rec}+L_{clu}
\end{equation}
The $L_{total}$ is directly employed to optimized the entire AMI-Net model.\\
\indent During the testing process, given a test image $I_{t}\in R^{H\times W\times C}$, AMI-Net employs a pre-trained CNN to extract multi-scale features $\mathcal{F}(I_t)$. Subsequently, the adaptive mask generator generates an adaptive mask for the features. Following this, the masked features are input into the inpainting network in order to obtain reconstruction features $\hat{\mathcal{F}}(I_t)$. Ultimately, the anomaly score map $A_s$ can be generated through the utilization of both the reconstructed features and the input features.
\begin{equation}
    A_s=\underbrace{\big\|\hat{\mathcal{F}}(I_t)-\mathcal{F}(I_t)\big\|}_{MSE}\times \underbrace{(1-\frac{\hat{\mathcal{F}}(I_t)\cdot\mathcal{F}(I_t)}{\big\|\hat{\mathcal{F}}(I_t)\big\|\times \big\|\mathcal{F}(I_t)\big\|})}_{cosine-distance}
\end{equation}
Where $A_s\in R^{H_F\times W_F}$. Then, we employ a bilinear upsampling operation to scale $A_s$ to the size of $H\times W$.
\subsection{Parameter Configuration}
The key hyperparameters of AMI-Net include $K$, $P$, $N_i$, $N_s$, and $\lambda$.\\
\indent The patch size $K$ influences the granularity of feature tokens. When $K$ is excessively large, it leads to the loss of fine details in the features. Conversely, when $K$ is too small, it results in a significant increase in computational workload, thereby reducing the inference speed. Therefore, in this study, we set $K$ to an appropriate value, specifically 4.\\
\indent The quantity of cluster tokens, indicated by $P$, significantly influences the learning process of clustering. When $P$ is too small, the model can only retain a limited number of typical normal features. Conversely, when $P$ is too large, the model struggles to fit effectively. Therefore, in this study, we designate $P$ as 8.\\
\indent The quantity of transformer blocks, referred to as $N_{i}$, within the inpainting network significantly influences the model's restoration capability. A small $N_{i}$ leads to inadequate restoration ability, rendering it difficult to effectively employ normal information for mending masked regions. Conversely, an excessively large $N_i$ introduces heightened computational complexity and the potential challenge of model fitting. Hence, for this study, we designate $N_i$ as 8.\\
\indent $N_s$ corresponds to the quantity of transformer blocks employed within the adaptive mask generator. With the escalation of $N_s$, there is a notable surge in computational complexity, accompanied by an increased susceptibility to introducing defective features into cluster tokens. Consequently, in this research, we opt to set $N_s$ at 1 to alleviate these potential challenges.\\
\indent $\lambda$ is employed to regulate the boundary size of clusters. When $\lambda$ is set excessively high, there is a risk of overlooking or missing defective features. Conversely, when $\lambda$ is set too low, normal features might be erroneously identified as anomalies. Therefore, in this study, we establish $\lambda$ at 0.5 to achieve an appropriate balance.
\begin{table*}[!t]
\centering
\caption{Anomaly detection and localization results in terms of image/pixel level AUROC on the MVTec AD dataset \cite{MVTEC}. $^\ddagger$ means our training with the feature jittering strategy proposed by reference \cite{uniad}.}
\label{table:mvtecad}
\begin{threeparttable}
\resizebox{\textwidth}{!}{
\begin{tabular}{c|c|cccccccc|>{\columncolor{green!5}}c>{\columncolor{blue!5}}c}
\Xhline{1.5pt}
{\color[HTML]{000000} } & {\color[HTML]{000000} Category}   & AE\_SSIM  & {\color[HTML]{000000} TrustMAE}  & {\color[HTML]{000000} RIAD}      & {\color[HTML]{000000} DFR}       & {\color[HTML]{000000} DRAEM}     & {\color[HTML]{000000} PatchCore} & {\color[HTML]{000000} MKD}       & {\color[HTML]{000000} MBPFM}     & {\color[HTML]{000000} \textbf{Ours}} & {\color[HTML]{000000} \textbf{Ours$^{\ddagger}$}}\\ \hline
{\color[HTML]{000000} } & {\color[HTML]{000000} Carpet}     & 67.0/87.0 & {\color[HTML]{000000} 97.4/98.5} & {\color[HTML]{000000} 84.2/94.2} & {\color[HTML]{000000} -/97.0}    & {\color[HTML]{000000} 97.0/95.5} & {\color[HTML]{000000} 98.7/\uline{99.0}} & {\color[HTML]{000000} 79.3/95.6} & {\color[HTML]{000000} \textbf{100}/\textbf{99.2}}  & {\color[HTML]{000000} \uline{99.8}/\textbf{99.2}} & {\color[HTML]{000000} \uline{99.8}/\textbf{99.2}}    \\
{\color[HTML]{000000} } & {\color[HTML]{000000} Grid}       & 69.0/94.0 & {\color[HTML]{000000} 99.1/97.5} & {\color[HTML]{000000} 99.6/96.3} & {\color[HTML]{000000} -/98.0}    & {\color[HTML]{000000} \uline{99.9}/\textbf{99.7}} & {\color[HTML]{000000} 98.2/98.7} & {\color[HTML]{000000} 78.0/91.8} & {\color[HTML]{000000} 98.0/98.8} & {\color[HTML]{000000} \textbf{100}/98.8}  & {\color[HTML]{000000} \uline{99.9}/\uline{98.9}}       \\
{\color[HTML]{000000} } & {\color[HTML]{000000} Leather}    & 46.0/78.0 & {\color[HTML]{000000} \uline{95.1}/98.1} & {\color[HTML]{000000} \textbf{100}/\textbf{99.4}}  & {\color[HTML]{000000} -/98.0}    & {\color[HTML]{000000} \textbf{100}/98.6}  & {\color[HTML]{000000} \textbf{100}/\uline{99.3}}  & {\color[HTML]{000000} \uline{95.1}/98.1} & {\color[HTML]{000000} \textbf{100}/\textbf{99.4}}  & {\color[HTML]{000000} \textbf{100}/\uline{99.3}}   & {\color[HTML]{000000} \textbf{100}/\textbf{99.4}}      \\
{\color[HTML]{000000} } & {\color[HTML]{000000} Tile}       & 52.0/59.0 & {\color[HTML]{000000} 97.3/82.5} & {\color[HTML]{000000} 98.7/89.1} & {\color[HTML]{000000} -/87.0}    & {\color[HTML]{000000} 99.6/\textbf{99.2}} & {\color[HTML]{000000} 98.7/95.6} & {\color[HTML]{000000} 91.6/82.8} & {\color[HTML]{000000} 99.6/\uline{96.2}} & {\color[HTML]{000000} \uline{99.9}/95.9}  & {\color[HTML]{000000} \textbf{100}/96.0}      \\
\multirow{-5}{*}{{\color[HTML]{000000} \rotatebox{90}{Texture}}} & {\color[HTML]{000000} Wood}       & 83.0/73.0 & {\color[HTML]{000000} \textbf{99.8}/92.6} & {\color[HTML]{000000} 93.0/85.8} & {\color[HTML]{000000} -/94.0}    & {\color[HTML]{000000} 99.1/\textbf{96.4}} & {\color[HTML]{000000} 99.2/95.0} & {\color[HTML]{000000} 94.3/84.8} & {\color[HTML]{000000} \uline{99.5}/\uline{95.6}} & {\color[HTML]{000000} 99.4/94.8} & {\color[HTML]{000000} 99.3/95.3}       \\ \hline
{\color[HTML]{000000} } & {\color[HTML]{000000} Average Texture}       & 63.4/78.2 & {\color[HTML]{000000} 97.7/93.8} & {\color[HTML]{000000} 95.1/93.9} & {\color[HTML]{000000} -/94.8}    & {\color[HTML]{000000} 99.1/\textbf{97.9}} & {\color[HTML]{000000} 99.0/97.5} & {\color[HTML]{000000} 87.7/90.6} & {\color[HTML]{000000} \uline{99.4}/\uline{97.8}} & {\color[HTML]{000000} \textbf{99.8}/97.6} & {\color[HTML]{000000} \textbf{99.8}/\uline{97.8}}       \\ \hline
{\color[HTML]{000000} } & {\color[HTML]{000000} Bottle}     & 88.0/93.0 & {\color[HTML]{000000} 97.0/93.4} & {\color[HTML]{000000} \uline{99.9}/98.4} & {\color[HTML]{000000} -/97.0}    & {\color[HTML]{000000} 99.2/99.1} & {\color[HTML]{000000} \textbf{100}/98.6}  & {\color[HTML]{000000} 99.4/96.3} & {\color[HTML]{000000} \textbf{100}/98.4}  & {\color[HTML]{000000} \textbf{100}/\uline{98.7}}   & {\color[HTML]{000000} \textbf{100}/\textbf{98.8}}      \\
{\color[HTML]{000000} } & {\color[HTML]{000000} Cable}      & 61.0/82.0 & {\color[HTML]{000000} 85.1/92.9} & {\color[HTML]{000000} 81.9/84.2} & {\color[HTML]{000000} -/92.0}    & {\color[HTML]{000000} 91.8/94.7} & {\color[HTML]{000000} \textbf{99.5}/\uline{98.4}} & {\color[HTML]{000000} 89.2/82.4} & {\color[HTML]{000000} 98.8/96.7} & {\color[HTML]{000000} \uline{99.1}/98.1}  & {\color[HTML]{000000} \textbf{99.5}/\textbf{98.6}}      \\
{\color[HTML]{000000} } & {\color[HTML]{000000} Capsule}    & 61.0/94.0 & {\color[HTML]{000000} 78.8/87.4} & {\color[HTML]{000000} 88.4/92.8} & {\color[HTML]{000000} -/\textbf{99.0}}    & {\color[HTML]{000000} \textbf{98.5}/94.3} & {\color[HTML]{000000} 98.1/98.8} & {\color[HTML]{000000} 80.5/95.9} & {\color[HTML]{000000} 94.5/98.3} & {\color[HTML]{000000} 95.7/98.6} & {\color[HTML]{000000} \uline{98.4}/\uline{98.9}}    \\
{\color[HTML]{000000} } & {\color[HTML]{000000} Hazelnut}   & 54.0/97.0 & {\color[HTML]{000000} 98.5/98.5} & {\color[HTML]{000000} 83.3/96.1} & {\color[HTML]{000000} -/99.0}    & {\color[HTML]{000000} \textbf{100}/\textbf{99.7}}  & {\color[HTML]{000000} \textbf{100}/98.7}  & {\color[HTML]{000000} 98.4/94.6} & {\color[HTML]{000000} \textbf{100}/\uline{99.1}}  & {\color[HTML]{000000} \uline{99.9}/98.3}  & {\color[HTML]{000000} \textbf{100}/98.6}      \\
{\color[HTML]{000000} } & {\color[HTML]{000000} Metal nut}  & 54.0/89.0 & {\color[HTML]{000000} 76.1/91.8} & {\color[HTML]{000000} 88.5/92.5} & {\color[HTML]{000000} -/93.0}    & {\color[HTML]{000000} 98.7/\textbf{99.5}} & {\color[HTML]{000000} \textbf{100}/\uline{98.4}}  & {\color[HTML]{000000} 73.6/86.4} & {\color[HTML]{000000} \textbf{100}/97.2}  & {\color[HTML]{000000} 99.2/95.3}  & {\color[HTML]{000000} \uline{99.8}/96.5}      \\
{\color[HTML]{000000} } & {\color[HTML]{000000} Pill}       & 60.0/91.0 & {\color[HTML]{000000} 83.3/89.9} & {\color[HTML]{000000} 83.8/95.7} & {\color[HTML]{000000} -/97.0}    & {\color[HTML]{000000} \textbf{98.9}/97.6} & {\color[HTML]{000000} \uline{96.6}/97.4} & {\color[HTML]{000000} 82.7/89.6} & {\color[HTML]{000000} 96.5/97.2} & {\color[HTML]{000000} 95.9/\uline{97.7}}  & {\color[HTML]{000000} 96.0/\textbf{98.4}}      \\
{\color[HTML]{000000} } & {\color[HTML]{000000} Screw}      & 51.0/96.0 & {\color[HTML]{000000} 83.4/97.6} & {\color[HTML]{000000} 84.5/98.8} & {\color[HTML]{000000} -/99.0}    & {\color[HTML]{000000} 93.9/97.6} & {\color[HTML]{000000} \textbf{98.1}/\textbf{99.4}} & {\color[HTML]{000000} 83.3/96.0} & {\color[HTML]{000000} 91.8/98.7} & {\color[HTML]{000000} 97.1/\uline{99.0}}  & {\color[HTML]{000000} \uline{97.9}/\textbf{99.4}}       \\
{\color[HTML]{000000} } & {\color[HTML]{000000} Toothbrush} & 74.0/92.0 & {\color[HTML]{000000} \uline{96.9}/98.1} & {\color[HTML]{000000} \textbf{100}/\textbf{98.9}}  & {\color[HTML]{000000} -/98.1}    & {\color[HTML]{000000} \textbf{100}/98.1}  & {\color[HTML]{000000} \textbf{100}/98.7}  & {\color[HTML]{000000} 92.2/96.1} & {\color[HTML]{000000} 88.6/98.6} & {\color[HTML]{000000} 93.6/\uline{98.8}}  & {\color[HTML]{000000} 96.1/\textbf{98.9}}      \\
{\color[HTML]{000000} } & {\color[HTML]{000000} Transistor} & 52.0/90.0 & {\color[HTML]{000000} 87.5/92.7} & {\color[HTML]{000000} 90.9/87.7} & {\color[HTML]{000000} -/80.0}    & {\color[HTML]{000000} 93.1/90.9} & {\color[HTML]{000000} \textbf{100}/96.3}  & {\color[HTML]{000000} 85.6/76.5} & {\color[HTML]{000000} \uline{97.8}/87.8} & {\color[HTML]{000000} \textbf{100}/\uline{96.7}}  & {\color[HTML]{000000} \textbf{100}/\textbf{98.2}}      \\
\multirow{-10}{*}{{\color[HTML]{000000} \rotatebox{90}{Object}}} & {\color[HTML]{000000} Zipper}     & 80.0/88.0 & {\color[HTML]{000000} 87.5/97.8} & {\color[HTML]{000000} 98.1/97.8} & {\color[HTML]{000000} -/96.0}    & {\color[HTML]{000000} \textbf{100}/\textbf{98.8}}  & {\color[HTML]{000000} \uline{99.4}/\uline{98.5}} & {\color[HTML]{000000} 93.2/93.9} & {\color[HTML]{000000} 97.4/98.2} & {\color[HTML]{000000} 97.8/98.2}  & {\color[HTML]{000000} 98.5/\uline{98.5}}      \\ \hline
{\color[HTML]{000000} } & {\color[HTML]{000000} Average Object}     & 63.5/91.2 & {\color[HTML]{000000} 87.4/94.0} & {\color[HTML]{000000} 89.9/94.3} & {\color[HTML]{000000} -/95.0}    & {\color[HTML]{000000} 97.4/97.0}  & {\color[HTML]{000000} \textbf{99.2}/\uline{98.3}} & {\color[HTML]{000000} 87.8/90.8} & {\color[HTML]{000000} 96.5/97.0} & {\color[HTML]{000000} 97.8/97.9}  & {\color[HTML]{000000} \uline{98.6}/\textbf{98.5}}      \\ \hline
{\color[HTML]{000000} } & {\color[HTML]{000000} Average All}    & 63.5/87.0 & {\color[HTML]{000000} 90.9/94.0} & {\color[HTML]{000000} 91.7/94.2} & {\color[HTML]{000000} 93.8/95.5} & {\color[HTML]{000000} 98.0/97.3} & {\color[HTML]{000000} \textbf{99.1}/\uline{98.1}} & {\color[HTML]{000000} 87.7/90.7} & {\color[HTML]{000000} 97.5/97.3} & {\color[HTML]{000000} 98.5/97.8} &  {\color[HTML]{000000} \uline{99.0}/\textbf{98.2}}     \\ \Xhline{1.5pt}
\end{tabular}}
\begin{tablenotes}
    \item[1] The best result is in \textbf{bold}, and the second best is \uline{underlined}.
\end{tablenotes}
\end{threeparttable}
\end{table*}
\section{Experiments}
In this section, we validate the detection performance of AMI-Net on publicly available MVTec AD \cite{MVTEC} and BTAD \cite{BTAD} datasets, while also assessing its inference speed. Furthermore, an extensive set of ablation experiments is conducted to showcase the influence of varying parameter configurations on the performance of AMI-Net.
\subsection{Experimental Configuration}
\subsubsection{Dataset}
We conducted performance validation on two benchmark datasets: MVTec AD and BTAD. The MVTec AD dataset consists of 15 categories of industrial defect images, including 5 categories of texture images and 10 categories of object images. This dataset contains 3629 normal images for training, and 498 normal images and 1982 defect images for testing. The BTAD dataset encompasses 3 categories of industrial texture images, presenting a more challenging pixel-level localization aspect. It comprises 1799 normal training images and 451 normal images along with 290 defect images for testing.
\subsubsection{Implementation Details}
The baseline model AMI-Net is trained over 200 epochs employing the AdamW optimizer with a learning rate of 0.001 and a batch size of 8. The enhanced model AMI-Net$^\ddagger$ incorporates a feature jittering strategy inspired by Reference \cite{uniad}. This strategy requires the model to effectively recover the source message, even when presented with inputs containing noise. Each image is resized to 256 × 256 and subjected to normalization using the mean and standard deviation extracted from the ImageNet dataset. All experimental runs are executed on a computer equipped with an Intel(R) Core(TM) i7-13700KF CPU operating at 3.40 GHz, coupled with an NVIDIA GeForce RTX 4090 GPU boasting a substantial 24GB memory capacity.
\begin{table}[!t]
\centering
\caption{Anomaly detection and localization results in terms of image/pixel level AUROC on the BTAD dataset \cite{BTAD}. The experimental results of other superior methods are sourced from references \cite{lei2023pyramidflow} and \cite{tien2023revisitingrd4ad}. $^\ddagger$ means our training with the feature jittering strategy proposed by reference \cite{uniad}.}
\label{table:btad}
\begin{threeparttable}
\resizebox{\linewidth}{!}{\begin{tabular}{ccccc}
\Xhline{1.5pt}
{\color[HTML]{000000} Methods}    & {\color[HTML]{000000} Class 01}  & {\color[HTML]{000000} Class 02}  & {\color[HTML]{000000} Class 03}  & {\color[HTML]{000000} Average}   \\ \hline
{\color[HTML]{000000} VT-ADL}     & {\color[HTML]{000000} 97.6/\textbf{99.0}} & {\color[HTML]{000000} 71.0/94.0} & {\color[HTML]{000000} 82.6/77.0} & {\color[HTML]{000000} 83.7/90.0} \\
{\color[HTML]{000000} Patch SVDD} & {\color[HTML]{000000} 95.7/91.6} & {\color[HTML]{000000} 72.1/93.6} & {\color[HTML]{000000} 82.1/91.0} & {\color[HTML]{000000} 83.3/92.1} \\
{\color[HTML]{000000} SPADE}      & {\color[HTML]{000000} 91.4/\uline{97.3}} & {\color[HTML]{000000} 71.4/94.4} & {\color[HTML]{000000} \uline{99.9}/99.1} & {\color[HTML]{000000} 87.6/96.9} \\
{\color[HTML]{000000} PatchCore}  & {\color[HTML]{000000} 90.9/95.5} & {\color[HTML]{000000} 79.3/94.7} & {\color[HTML]{000000} 99.8/99.3} & {\color[HTML]{000000} 90.0/96.5} \\
{\color[HTML]{000000} FastFlow}   & {\color[HTML]{000000} \uline{99.4}/97.1} & {\color[HTML]{000000} 82.4/93.6} & {\color[HTML]{000000} 91.1/98.3} & {\color[HTML]{000000} 90.1/96.3} \\
{\color[HTML]{000000} CFA}        & {\color[HTML]{000000} 98.1/95.9} & {\color[HTML]{000000} \textbf{85.5}/\textbf{96.0}} & {\color[HTML]{000000} 99.0/98.6} & {\color[HTML]{000000} \uline{94.2}/96.8} \\ \hline
\cellcolor{green!5}\textbf{Ours} & \cellcolor{green!5}\textbf{99.9}/96.7  & \cellcolor{green!5}\uline{85.4}/\uline{95.2}  & \cellcolor{green!5}\textbf{100}/\uline{99.5}  & \cellcolor{green!5}\textbf{95.1}/\uline{97.1}  \\ 
\cellcolor{blue!5}\textbf{Ours$^\ddagger$} & \cellcolor{blue!5}\textbf{99.9}/96.8  & \cellcolor{blue!5}\uline{85.4}/\textbf{96.0}  & \cellcolor{blue!5}\textbf{100}/\textbf{99.6}  & \cellcolor{blue!5}\textbf{95.1}/\textbf{97.5}\\ 
\Xhline{1.5pt}
\end{tabular}}
\begin{tablenotes}
\footnotesize  
    \item[1] The best result is in \textbf{bold}, and the second best is \uline{underlined}.
\end{tablenotes}
\end{threeparttable}
\end{table}
\subsubsection{Evaluation Criteria}
We adopt widely employed Area Under the Receiver Operating Characteristic Curve (AUROC) metrics at both the image and pixel levels as the standard for defect detection and localization. Additionally, in the ablation experiments, we incorporate Average Precision (AP) metrics at both image and pixel levels.
\subsubsection{Comparative Methods}
In our experiments, we compare our proposed AMI-Net with the following outstanding methods: AE\_SSIM \cite{AE-SSIM}, TrustMAE \cite{TrustMAE}, RIAD \cite{RIAD}, DFR \cite{DFR}, DRAEM \cite{draem}, PatchCore \cite{PacthCore}, MKD \cite{MKD}, MBPFM \cite{MBPFM}, VT-ADL \cite{BTAD}, Patch SVDD \cite{PatchSVDD}, SPADE \cite{SPADE}, FastFlow \cite{fastflow}, CFA \cite{cfa}, US \cite{ST}, PaDiM \cite{PaDiM}, SimpleNet \cite{liu2023simplenet}, UniAD \cite{uniad}, TDG \cite{TDG}, DiffNet \cite{DiffNet}, and RegAD \cite{RegAD}.\\
\indent AE\_SSIM enhances detection performance by considering the structural information of images. TrustMAE employs a memory module within trust regions to reconstruct images and uses perceptual distance for anomaly detection and localization. RIAD uses a masking strategy and an image inpainting network to transform the reconstruction task into an image inpainting task. DRAEM proposes using natural images to synthesize artificial defect images and then employs a U-Net to predict and locate defect areas. Patch SVDD can establish multiple normal clusters at the patch level, enabling precise anomaly localization. PaDiM models normal features using multivariate Gaussian distributions, and calculates anomaly scores by measuring the Mahalanobis distance between the test features and the Gaussian distribution. SPADE, CFA, and PatchCore employ a memory bank to store normal features and determine anomalies based on the distance between test features and the most similar normal feature in the memory bank. MKD and US employ knowledge distillation for anomaly detection. MBPFM utilizes a multi-level bidirectional feature mapping for precise anomaly localization. FastFlow employs probability distributions to model normal data and detects anomalies by assessing deviations from the learned distribution. VT-ADL utilizes a transformer network to preserve spatial information of embedding patches, followed by the utilization of a Gaussian mixture density network for anomaly region localization. Simplenet introduces a feature discriminator for defect localization. UniAD proposes a neighbor masked attention and a layer-wise query decoder to achieve multi-class anomaly detection. TDG presents a hierarchical generative model that captures the multi-scale patch distribution in support images. DiffNet employs pre-trained features to estimate density via a normalizing flow. RegAD introduces a novel few-shot anomaly detection paradigm, employing meta-learning to learn a shared common model across multiple categories, with the goal of minimizing redundancy in anomaly detection tasks.
\subsection{Overall Performance Comparative Experiment}
\subsubsection{Experiment on MVTec AD Dataset}
\begin{figure*}[!t]
    \centering
    \includegraphics[width=185mm]{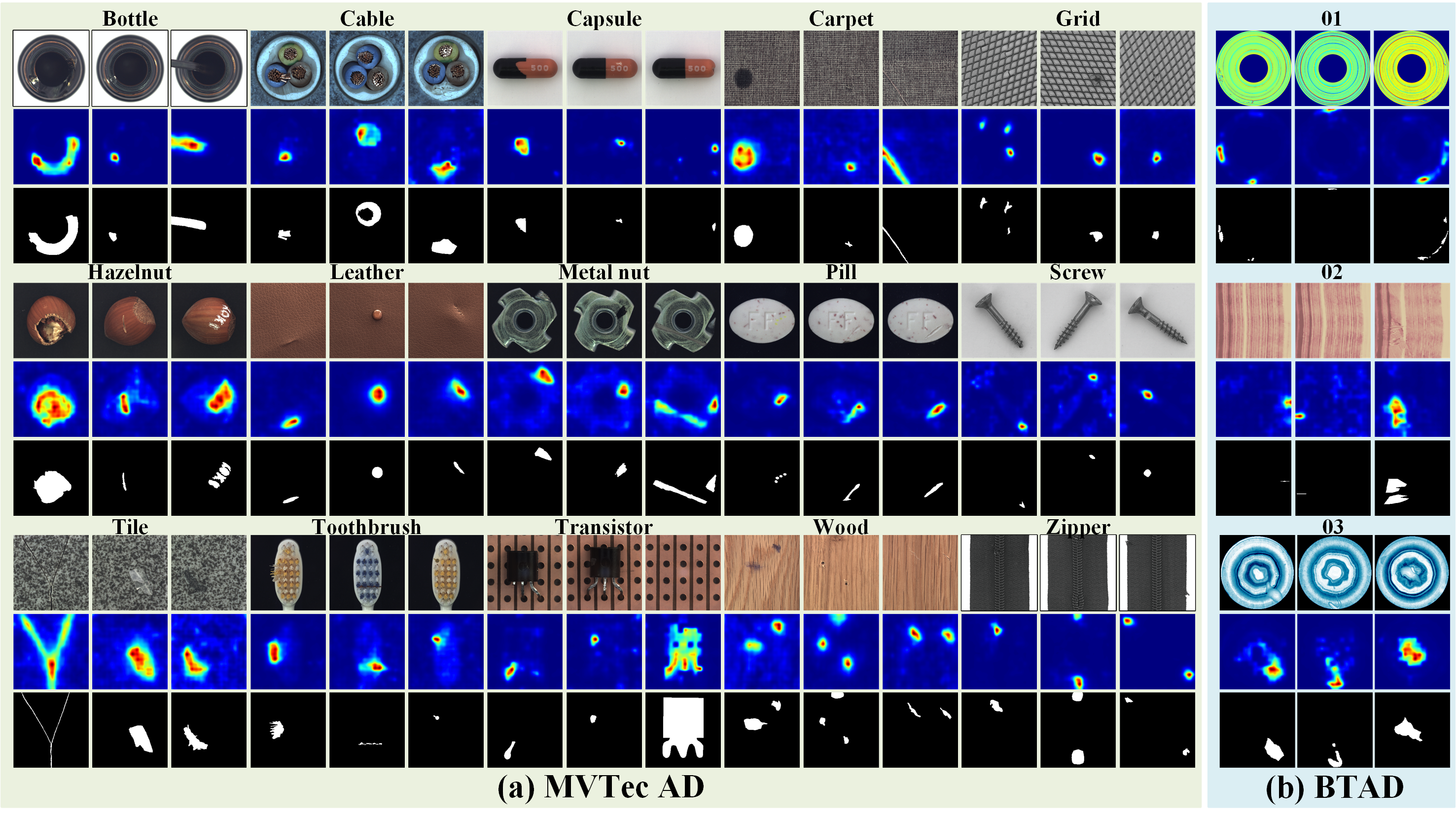}
    \caption{Localization Results of AMI-Net on MVTec AD \cite{MVTEC} and BTAD \cite{BTAD}. For each set, from top to bottom, there are the defect image, detection heat map, and corresponding label. (a) Competitive detection results across 15 categories in MVTec AD. (b) Competitive detection results across 3 categories in BTAD.}
    \label{fig:07}
\end{figure*}

To validate the detection performance of AMI-Net, we compare it against several state-of-the-art anomaly detection methods on the MVTec AD \cite{MVTEC} dataset, including AE\_SSIM \cite{AE-SSIM}, TrsutMAE \cite{TrustMAE}, RIAD \cite{RIAD}, DFR \cite{DFR}, DRAEM \cite{draem}, PatchCore \cite{PacthCore}, MKD \cite{MKD}, and MBPFM \cite{MBPFM}.\\
\indent Table \ref{table:mvtecad} presents the quantitative experimental results. AMI-Net (\textbf{Ours}$^\ddagger$ in the table) achieves the second-best detection performance, trailing only behind PatchCore, and it attains the best localization. Notably, AMI-Net attains a perfect image-level AUROC of 100\% on the Leather, Wood, Bottle, Hazelnut and Transistor datasets. Compared to the existing mask-based method RIAD, AMI-Net achieves a notable improvement of +7.3/+4.0\% in image/pixel-level AUROC respectively across the 15 categories. This enhancement can be attributed to the introduced adaptive mask generator, which generates masks during the testing process that obscure defect areas while retaining normal regions. This approach significantly bolsters the model's detection performance.\\
\indent Qualitative detection results of AMI-Net are depicted in Fig. \ref{fig:07}(a). AMI-Net accurately localizes defect regions across all 15 categories of data. It is noteworthy that global defects such as ``cable swap'' in the Cable category and ``misplaced" defects in the Transistor category are effectively detected by AMI-Net, attributed to the inclusion of position encoding during clustering.
\subsubsection{Experiment on BTAD Dataset}
To further validate the performance of AMI-Net, we conduct comparative evaluations on the challenging BTAD \cite{BTAD} dataset against several state-of-the-art approaches, including VT-ADL \cite{BTAD}, Patch SVDD \cite{PatchSVDD}, SPADE \cite{SPADE}, PatchCore \cite{PacthCore}, FastFlow \cite{fastflow}, and CFA \cite{cfa}.\\
\indent Quantitative experimental results are presented in Table \ref{table:btad}. AMI-Net (\textbf{Ours}$^\ddagger$ in the table) achieves the best detection and localization outcomes. In terms of the image/pixel AUROC metric, AMI-Net outperformes the second-best result by +0.9/+0.7\%, respectively. Notably, AMI-Net attains a 100\% image-wise AUROC on the class 03 dataset, further confirming the superiority of AMI-Net's detection capabilities.\\
\indent The \begin{figure}[!t]
    \centering
    \includegraphics[width=88mm]{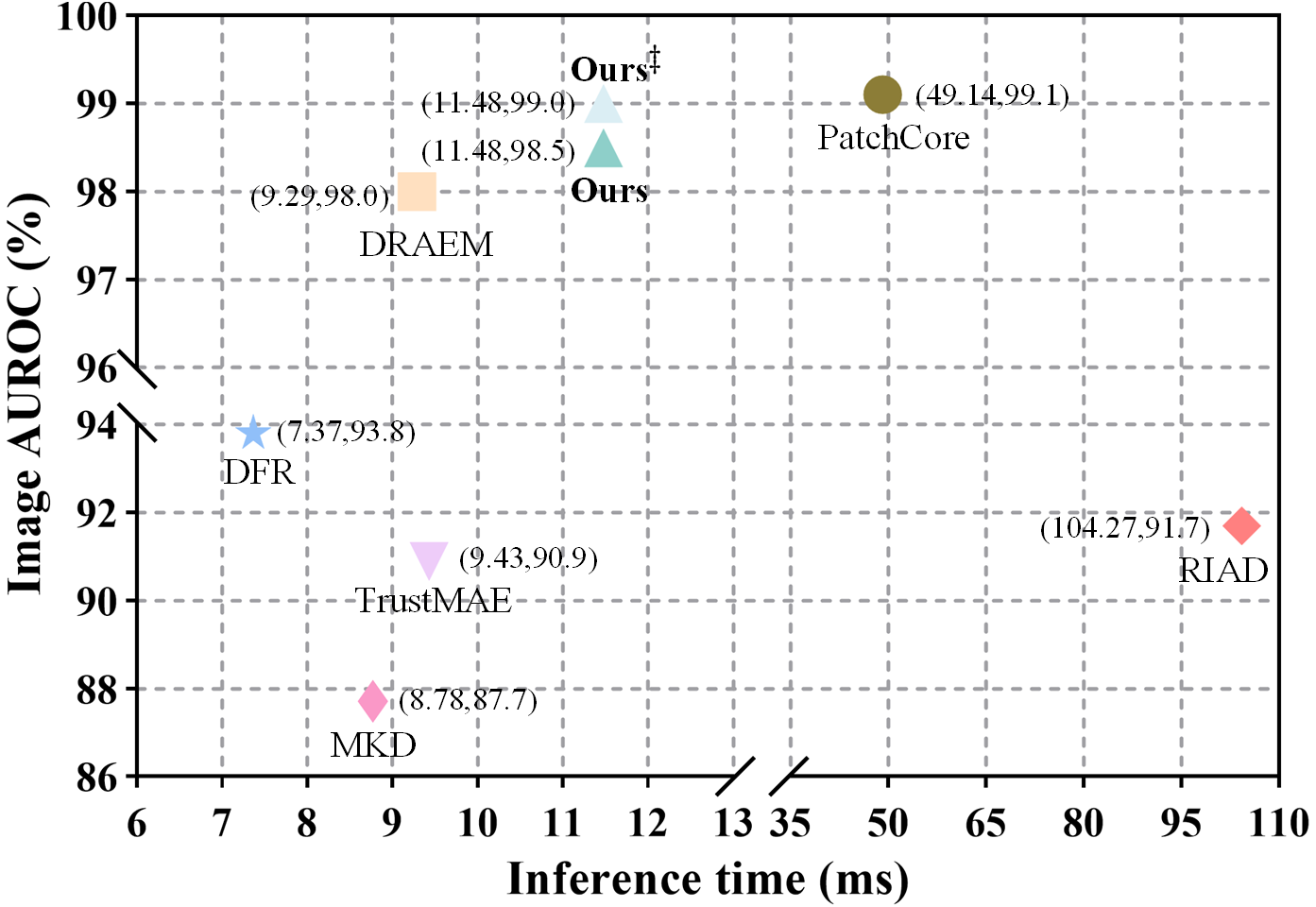}
    \caption{Inference time versus Image level AUROC on MVTec AD dataset \cite{MVTEC}. ($\cdot$, $\cdot$) denotes (Inference time, Image AUROC).}
    \label{fig:08}
    \vspace{-1.5em}
\end{figure}qualitative detection results of AMI-Net are depicted in Fig. \ref{fig:07}(b). AMI-Net demonstrates the capability to accurately locate even very small defects, achieving precise localization across all three datasets. This affirms its detection generalization prowess.
\subsubsection{Potential reasons why AMI-Net is not the best in MVTec AD but performs best in BTAD}
During the training process, the adaptive mask generator of AMI-Net requires clustering operations on the input features. In comparison to texture categories, object categories exhibit greater pose diversity, posing a more formidable challenge in the clustering process and subsequently leading to a performance decline. Therefore, AMI-Net is more suitable for the detection of texture categories. The MVTec AD dataset comprises 5 texture and 10 object categories, whereas the BTAD dataset includes only 3 texture categories. Consequently, on the BTAD dataset, AMI-Net performs exceptionally well, whereas it does not achieve optimal results on the MVTec AD dataset.

\subsubsection{Real-Time Analysis}
Industrial anomaly detection necessitates a favorable trade-off between detection accuracy and speed. Hence, we conduct a real-time analysis of AMI-Net and other competitive methods.\\
\indent The quantitative real-time analysis results are illustrated in Fig. \ref{fig:08}. While PatchCore achieves the highest image AUROC (99.1\%) on the MVTec AD dataset, its inference time is excessively long at 49.14ms. In contrast, our proposed method exhibits the second-best performance (99.0\%) with an inference time of 11.48ms. Although our proposed method lags behind PatchCore by only 0.1\% in detection accuracy, its inference speed is over \textbf{$4 \times$} faster than PatchCore. Additionally, as shown in Table \ref{table:btad}, on the BTAD dataset, PatchCore achieves only a 90.0\% image AUROC and 96.5\% pixel AUROC, whereas our AMI-Net achieves a 95.1\% image AUROC and 97.5\% pixel AUROC. This indicates that AMI-Net possesses better generalization compared to PatchCore, maintaining superior detection performance across different datasets. Compared to the existing mask-based method RIAD, AMI-Net surpasses RIAD in both detection speed and accuracy. Although DRAEM, DFR, TrustMAE, and MKD exhibit superior detection speeds compared to AMI-Net, they significantly lag behind in detection accuracy. Consequently, taking a comprehensive view, AMI-Net achieves a commendable balance between detection accuracy and speed.
\begin{table*}[!t]
\centering
\caption{Anomaly detection and localization results in terms of image/pixel level AUROC on the MVTec AD dataset \cite{MVTEC}. All methods are evaluated under the unified (one for all) case. In the unified case, the learned model is applied to detect anomalies for all categories without fine-tuning. The experimental results of other superior methods are sourced from reference \cite{lu2023hierarchical}. $^\ddagger$ means our training with the feature jittering strategy proposed by reference \cite{uniad}.}
\label{table:mvtecad_unified}
\begin{threeparttable}
\resizebox{\textwidth}{!}{
\begin{tabular}{c|c|cccccccc|>{\columncolor{green!5}}c>{\columncolor{blue!5}}c}
\Xhline{1.5pt}
{\color[HTML]{000000} } & {\color[HTML]{000000} Category}   & {\color[HTML]{000000} US}        & {\color[HTML]{000000} Patch SVDD}     & {\color[HTML]{000000} PaDiM}     & {\color[HTML]{000000} MKD}       & {\color[HTML]{000000} DRAEM}     & {\color[HTML]{000000} SimpleNet} & {\color[HTML]{000000} PatchCore} & {\color[HTML]{000000} UniAD}     & {\color[HTML]{000000} \textbf{Ours}}   & {\color[HTML]{000000} \textbf{Ours$^\ddagger$}}  \\ \hline
{\color[HTML]{000000} } & {\color[HTML]{000000} Carpet}     & {\color[HTML]{000000} 86.6/88.7} & {\color[HTML]{000000} 63.3/78.6} & {\color[HTML]{000000} 93.8/97.6} & {\color[HTML]{000000} 69.8/95.5} & {\color[HTML]{000000} \uline{98.0}/\uline{98.6}} & {\color[HTML]{000000} 95.9/92.4} & {\color[HTML]{000000} 97.0/98.1} & {\color[HTML]{000000} \textbf{99.8}/\uline{98.5}} & {\color[HTML]{000000} 97.5/98.2} & {\color[HTML]{000000} 97.9/\textbf{98.6}} \\
{\color[HTML]{000000} } & {\color[HTML]{000000} Grid}       & {\color[HTML]{000000} 69.2/64.5} & {\color[HTML]{000000} 66.0/70.8} & {\color[HTML]{000000} 73.9/71.0} & {\color[HTML]{000000} 83.8/82.3} & {\color[HTML]{000000} \textbf{99.3}/\textbf{98.7}} & {\color[HTML]{000000} 49.8/46.7} & {\color[HTML]{000000} 91.4/98.4} & {\color[HTML]{000000} 98.2/96.5} & {\color[HTML]{000000} 95.1/94.1} & {\color[HTML]{000000} \uline{99.2}/\uline{98.5}} \\
{\color[HTML]{000000} } & {\color[HTML]{000000} Leather}    & {\color[HTML]{000000} 97.2/95.4} & {\color[HTML]{000000} 60.8/93.5} & {\color[HTML]{000000} \uline{99.9}/84.8} & {\color[HTML]{000000} 93.6/96.7} & {\color[HTML]{000000} 98.7/97.3} & {\color[HTML]{000000} 93.9/96.9} & {\color[HTML]{000000} \textbf{100}/\textbf{99.2}}  & {\color[HTML]{000000} \textbf{100}/98.8}  & {\color[HTML]{000000} \textbf{100}/\uline{98.9}}  & {\color[HTML]{000000} \textbf{100}/98.8}  \\
{\color[HTML]{000000} } & {\color[HTML]{000000} Tile}       & {\color[HTML]{000000} 93.7/82.7} & {\color[HTML]{000000} 88.3/92.1} & {\color[HTML]{000000} 93.3/80.5} & {\color[HTML]{000000} 89.5/85.3} & {\color[HTML]{000000} \textbf{99.8}/\textbf{98.0}} & {\color[HTML]{000000} 93.7/93.1} & {\color[HTML]{000000} 96.0/90.3} & {\color[HTML]{000000} 99.3/91.8} & {\color[HTML]{000000} 98.7/91.9} & {\color[HTML]{000000} \uline{99.5}/\uline{94.3}} \\ 
 \multirow{-5}{*}{{\color[HTML]{000000} \rotatebox{90}{Texture}}} & {\color[HTML]{000000} Wood}       & {\color[HTML]{000000} 90.6/83.3} & {\color[HTML]{000000} 72.1/80.7} & {\color[HTML]{000000} 98.4/89.1} & {\color[HTML]{000000} 93.4/80.5} & {\color[HTML]{000000} \uline{99.8}/\textbf{96.0}} & {\color[HTML]{000000} 95.2/84.8} & {\color[HTML]{000000} 93.8/90.8} & {\color[HTML]{000000} 98.6/93.2} & {\color[HTML]{000000} 99.6/92.1} & {\color[HTML]{000000} \textbf{100}/\uline{93.3}}  \\ \hline
% {\color[HTML]{000000} Average texture}       & {\color[HTML]{000000} 90.6/83.3} & {\color[HTML]{000000} 72.1/80.7} & {\color[HTML]{000000} 98.4/89.1} & {\color[HTML]{000000} 93.4/80.5} & {\color[HTML]{000000} \uline{99.8}/\textbf{96.0}} & {\color[HTML]{000000} 95.2/84.8} & {\color[HTML]{000000} 93.8/90.8} & {\color[HTML]{000000} 98.6/93.2} & {\color[HTML]{000000} 99.6/92.1} & {\color[HTML]{000000} \textbf{100}/\uline{93.3}}  \\ \hline
{\color[HTML]{000000} }   & {\color[HTML]{000000} Average Texture}    & {\color[HTML]{000000} 87.5/82.9}   & {\color[HTML]{000000} 70.1/83.1}  & {\color[HTML]{000000} 91.9/84.6}     & {\color[HTML]{000000} 86.0/88.1}      & {\color[HTML]{000000} 99.1/\textbf{97.7}}               & {\color[HTML]{000000} 85.7/82.8}    & {\color[HTML]{000000} 95.6/95.4}   & {\color[HTML]{000000} \uline{99.2}/95.8}     & {\color[HTML]{000000} 98.2/95.0}     & {\color[HTML]{000000} \textbf{99.3}/\uline{96.7}}             \\ \hline
{\color[HTML]{000000} } & {\color[HTML]{000000} Bottle}     & {\color[HTML]{000000} 84.0/67.9} & {\color[HTML]{000000} 85.5/86.7} & {\color[HTML]{000000} 97.9/96.1} & {\color[HTML]{000000} 98.7/91.8} & {\color[HTML]{000000} 97.5/87.6} & {\color[HTML]{000000} 97.7/91.2} & {\color[HTML]{000000} \textbf{100}/97.4}  & {\color[HTML]{000000} \uline{99.7}/\uline{98.1}} & {\color[HTML]{000000} \textbf{100}/97.8}  & {\color[HTML]{000000} \textbf{100}/\textbf{98.4}}  \\
{\color[HTML]{000000} } & {\color[HTML]{000000} Cable}      & {\color[HTML]{000000} 60.0/78.3} & {\color[HTML]{000000} 64.4/62.2} & {\color[HTML]{000000} 70.9/81.0} & {\color[HTML]{000000} 78.2/89.3} & {\color[HTML]{000000} 57.8/71.3} & {\color[HTML]{000000} 87.6/88.1} & {\color[HTML]{000000} 95.3/93.6} & {\color[HTML]{000000} 95.2/\uline{97.3}} & {\color[HTML]{000000} \uline{96.6}/94.9} & {\color[HTML]{000000} \textbf{98.7}/\textbf{98.2}} \\
{\color[HTML]{000000} } & {\color[HTML]{000000} Capsule}    & {\color[HTML]{000000} 57.6/85.5} & {\color[HTML]{000000} 61.3/83.1} & {\color[HTML]{000000} 73.4/96.9} & {\color[HTML]{000000} 68.3/88.3} & {\color[HTML]{000000} 65.3/50.5} & {\color[HTML]{000000} 78.3/89.7} & {\color[HTML]{000000} \textbf{96.8}/98.0} & {\color[HTML]{000000} 86.9/\uline{98.5}} & {\color[HTML]{000000} 83.1/98.2} & {\color[HTML]{000000} \uline{94.1}/\textbf{98.9}} \\
{\color[HTML]{000000} } & {\color[HTML]{000000} Hazelnut}   & {\color[HTML]{000000} 95.8/93.7} & {\color[HTML]{000000} 83.9/97.4} & {\color[HTML]{000000} 85.5/96.3} & {\color[HTML]{000000} 97.1/91.2} & {\color[HTML]{000000} 93.7/96.9} & {\color[HTML]{000000} 99.2/95.7} & {\color[HTML]{000000} 99.3/97.6} & {\color[HTML]{000000} \textbf{99.8}/\textbf{98.1}} & {\color[HTML]{000000} \uline{99.6}/97.0} & {\color[HTML]{000000} \textbf{99.8}/\uline{98.0}} \\
{\color[HTML]{000000} } & {\color[HTML]{000000} Metal nut}  & {\color[HTML]{000000} 62.7/76.6} & {\color[HTML]{000000} 80.9/\uline{96.0}} & {\color[HTML]{000000} 88.0/84.8} & {\color[HTML]{000000} 64.9/64.2} & {\color[HTML]{000000} 72.8/62.2} & {\color[HTML]{000000} 85.1/90.9} & {\color[HTML]{000000} \uline{99.1}/\textbf{96.3}} & {\color[HTML]{000000} \textbf{99.2}/94.8} & {\color[HTML]{000000} 98.6/92.3} & {\color[HTML]{000000} 98.5/\uline{96.0}} \\
{\color[HTML]{000000} } & {\color[HTML]{000000} Pill}       & {\color[HTML]{000000} 56.1/80.3} & {\color[HTML]{000000} 89.4/\uline{96.5}} & {\color[HTML]{000000} 68.8/87.7} & {\color[HTML]{000000} 79.7/69.7} & {\color[HTML]{000000} 82.2/94.4} & {\color[HTML]{000000} 78.3/89.7} & {\color[HTML]{000000} 86.4/90.8} & {\color[HTML]{000000} \textbf{93.7}/95.0} & {\color[HTML]{000000} 91.9/95.4} & {\color[HTML]{000000} \uline{93.6}/\textbf{97.4}} \\
{\color[HTML]{000000} } & {\color[HTML]{000000} Screw}      & {\color[HTML]{000000} 66.9/90.8} & {\color[HTML]{000000} 80.9/74.3} & {\color[HTML]{000000} 56.9/94.1} & {\color[HTML]{000000} 75.6/92.1} & {\color[HTML]{000000} \uline{92.0}/95.5} & {\color[HTML]{000000} 45.5/93.7} & {\color[HTML]{000000} \textbf{94.2}/\textbf{98.9}} & {\color[HTML]{000000} 87.5/\uline{98.3}} & {\color[HTML]{000000} 80.0/97.1} & {\color[HTML]{000000} 83.7/\textbf{98.9}} \\
{\color[HTML]{000000} } & {\color[HTML]{000000} Toothbrush} & {\color[HTML]{000000} 57.8/86.9} & {\color[HTML]{000000} \uline{99.4}/98.0} & {\color[HTML]{000000} 95.3/95.6} & {\color[HTML]{000000} 75.3/88.9} & {\color[HTML]{000000} 90.6/97.7} & {\color[HTML]{000000} 94.7/97.5} & {\color[HTML]{000000} \textbf{100}/\textbf{98.8}}  & {\color[HTML]{000000} 94.2/98.4} & {\color[HTML]{000000} 97.2/97.7} & {\color[HTML]{000000} 95.3/\uline{98.7}} \\
{\color[HTML]{000000} } & {\color[HTML]{000000} Transistor} & {\color[HTML]{000000} 61.0/68.3} & {\color[HTML]{000000} 77.5/78.5} & {\color[HTML]{000000} 86.6/92.3} & {\color[HTML]{000000} 73.4/71.7} & {\color[HTML]{000000} 74.8/64.5} & {\color[HTML]{000000} 82.0/86.0} & {\color[HTML]{000000} 98.9/92.3} & {\color[HTML]{000000} \textbf{99.8}/\textbf{97.9}} & {\color[HTML]{000000} 98.6/90.5} & {\color[HTML]{000000} \uline{99.5}/\uline{95.5}} \\
\multirow{-10}{*}{{\color[HTML]{000000} \rotatebox{90}{Object}}} & {\color[HTML]{000000} Zipper}     & {\color[HTML]{000000} 78.6/84.2} & {\color[HTML]{000000} 77.8/95.1} & {\color[HTML]{000000} 79.7/94.8} & {\color[HTML]{000000} 87.4/86.1} & {\color[HTML]{000000} \uline{98.8}/\textbf{98.3}} & {\color[HTML]{000000} \textbf{99.1}/97.0} & {\color[HTML]{000000} 97.1/95.7} & {\color[HTML]{000000} 95.8/96.8} & {\color[HTML]{000000} 97.7/\uline{97.8}} & {\color[HTML]{000000} 97.5/\textbf{98.3}} \\ \hline
% {\color[HTML]{000000} Average object}     & {\color[HTML]{000000} 78.6/84.2} & {\color[HTML]{000000} 77.8/95.1} & {\color[HTML]{000000} 79.7/94.8} & {\color[HTML]{000000} 87.4/86.1} & {\color[HTML]{000000} \uline{98.8}/\textbf{98.3}} & {\color[HTML]{000000} \textbf{99.1}/97.0} & {\color[HTML]{000000} 97.1/95.7} & {\color[HTML]{000000} 95.8/96.8} & {\color[HTML]{000000} 97.7/\uline{97.8}} & {\color[HTML]{000000} 97.5/\textbf{98.3}} \\ \hline
 {\color[HTML]{000000} } & {\color[HTML]{000000} Average Object}  & {\color[HTML]{000000} 68.1/81.3} & {\color[HTML]{000000} 80.1/86.8} & {\color[HTML]{000000} 80.3/92.0} & {\color[HTML]{000000} 79.9/83.3} & {\color[HTML]{000000} 82.6/81.9} & {\color[HTML]{000000} 84.8/92.0} & {\color[HTML]{000000} \textbf{96.7}/95.9} & {\color[HTML]{000000} 95.2/\uline{97.3}}                      & {\color[HTML]{000000} 94.3/95.8} & {\color[HTML]{000000} \uline{96.1}/\textbf{97.8}} \\ \hline
{\color[HTML]{000000} } & {\color[HTML]{000000} Average All}    & {\color[HTML]{000000} 74.5/81.8} & {\color[HTML]{000000} 76.8/85.6} & {\color[HTML]{000000} 84.2/89.5} & {\color[HTML]{000000} 81.9/84.9} & {\color[HTML]{000000} 88.1/87.2} & {\color[HTML]{000000} 85.1/88.9} & {\color[HTML]{000000} 96.4/95.7} & {\color[HTML]{000000} \uline{96.5}/\uline{96.8}} & {\color[HTML]{000000} 95.6/95.6} & {\color[HTML]{000000} \textbf{97.2}/\textbf{97.5}} \\ \Xhline{1.5pt}
\end{tabular}}
\begin{tablenotes}
    \item[1] The best result is in \textbf{bold}, and the second best is \uline{underlined}.
\end{tablenotes}
\end{threeparttable}
\end{table*}
\subsection{Multi-Class Anomaly Detection with A Unified Model}
Traditional anomaly detection methods  typically follow a one-for-one scheme, training separate models for different categories of objects, which is time-consuming for practical industrial applications. Recently, UniAD \cite{uniad} proposed a one-for-all scheme, aiming to utilize a unified model for detecting anomalies across all different object categories without any fine-tuning, which is more challenging than traditional anomaly detection because it requires accurate modeling of multi-class distribution. Consequently, we compare the proposed AMI-Net with several outstanding methods under the unified setting on MVTec AD dataset \cite{MVTEC}, including US \cite{ST}, Patch SVDD \cite{PatchSVDD}, PaDiM \cite{PaDiM}, MKD \cite{MKD}, DRAEM \cite{draem}, SimpleNet \cite{liu2023simplenet}, PatchCore \cite{PacthCore}, and UniAD \cite{uniad}.\\
\indent Quantitative experimental results are presented in Table \ref{table:mvtecad_unified}. AMI-Net (\textbf{Ours}$^\ddagger$ in the table) achieves the best overall detection and localization performance with image/pixel level AUROC of 97.2/97.5\%, respectively, surpassing the current state-of-the-art method UniAD by +0.7/+0.7\%. Notably, AMI-Net achieves 100\% image level AUROC on the Leather, Wood, and Bottle datasets. This strongly underscores the superiority of AMI-Net, showcasing not only its outstanding detection performance in the traditional one-for-one setting but also its ability to maintain excellent performance in the more challenging one-for-all setting. This highlights its significant potential in industrial scenarios.

\begin{table}[!t]
\centering
\caption{Image/pixel level AUROC of k-shot anomaly detection on the MVTec AD dataset \cite{MVTEC}. The results are averaged over all categories. $^\ddagger$ means our training with the feature jittering strategy proposed by reference \cite{uniad}.}
\label{table:mvtecad_fewshot}
\begin{threeparttable}
\begin{tabular}{c|c|ccc|>{\columncolor{green!5}}c>{\columncolor{blue!5}}c}
\Xhline{1.5pt}
{\color[HTML]{000000} Dataset}                 & {\color[HTML]{000000} k} & {\color[HTML]{000000} TDG}    & {\color[HTML]{000000} DiffNet} & {\color[HTML]{000000} RegAD}     & {\color[HTML]{000000} \textbf{Ours}}      & {\color[HTML]{000000} \textbf{Ours$^\ddagger$}}     \\ \hline
{\color[HTML]{000000} }                        & {\color[HTML]{000000} 2} & {\color[HTML]{000000} 71.2/-} & {\color[HTML]{000000} 80.6/-}  & {\color[HTML]{000000} \textbf{85.7}/\uline{94.6}} & {\color[HTML]{000000} 84.2/94.4} & {\color[HTML]{000000} \uline{84.6}/\textbf{94.8}} \\
{\color[HTML]{000000} }                        & {\color[HTML]{000000} 4} & {\color[HTML]{000000} 72.7/-} & {\color[HTML]{000000} 81.3/-}  & {\color[HTML]{000000} \textbf{88.2}/\textbf{95.8}} & {\color[HTML]{000000} 86.7/95.5} & {\color[HTML]{000000} \uline{86.9}/\textbf{95.8}} \\
\multirow{-3}{*}{{\color[HTML]{000000} MVTec}} & {\color[HTML]{000000} 8} & {\color[HTML]{000000} 75.2/-} & {\color[HTML]{000000} 82.3-}   & {\color[HTML]{000000} \textbf{91.2}/\uline{96.7}} & {\color[HTML]{000000} \uline{90.9}/96.6} & {\color[HTML]{000000} \uline{90.9}/\textbf{96.8}} \\ \Xhline{1.5pt}
\end{tabular}
\begin{tablenotes}
     \item[1] The best result is in \textbf{bold}, and the second best is \uline{underlined}.
\end{tablenotes}
\end{threeparttable}
\end{table}
\subsection{Few Shot Anomaly Detection}
In real industrial scenarios, there may be situations where the number of samples is limited, and maintaining good performance under such circumstances is crucial. Therefore, we evaluate AMI-Net in the few-shot setting on MVTec AD dataset \cite{MVTEC} and compare its performance with some algorithms specifically designed for few-shot anomaly detection, including TDG \cite{TDG}, DiffNet \cite{DiffNet}, and RegAD \cite{RegAD}.\\
\indent Quantitative experimental results are illustrated in Table \ref{table:mvtecad_fewshot}. AMI-Net (\textbf{Ours$^\ddagger$} in the table) achieves the best localization results and the second-best detection performance, trailing only slightly behind RegAD. It is noteworthy that RegAD is a method specifically designed for few-shot anomaly detection, and its performance in other settings is not outstanding. In contrast, our AMI-Net not only demonstrates commendable performance in the few-shot setting but also showcases excellent detection performance in other settings such as one-for-one and one-for-all. This highlights the exceptional generalization ability of AMI-Net.
\subsection{Ablation Experiments on MVTec AD}
\begin{figure*}
    \centering
    \includegraphics{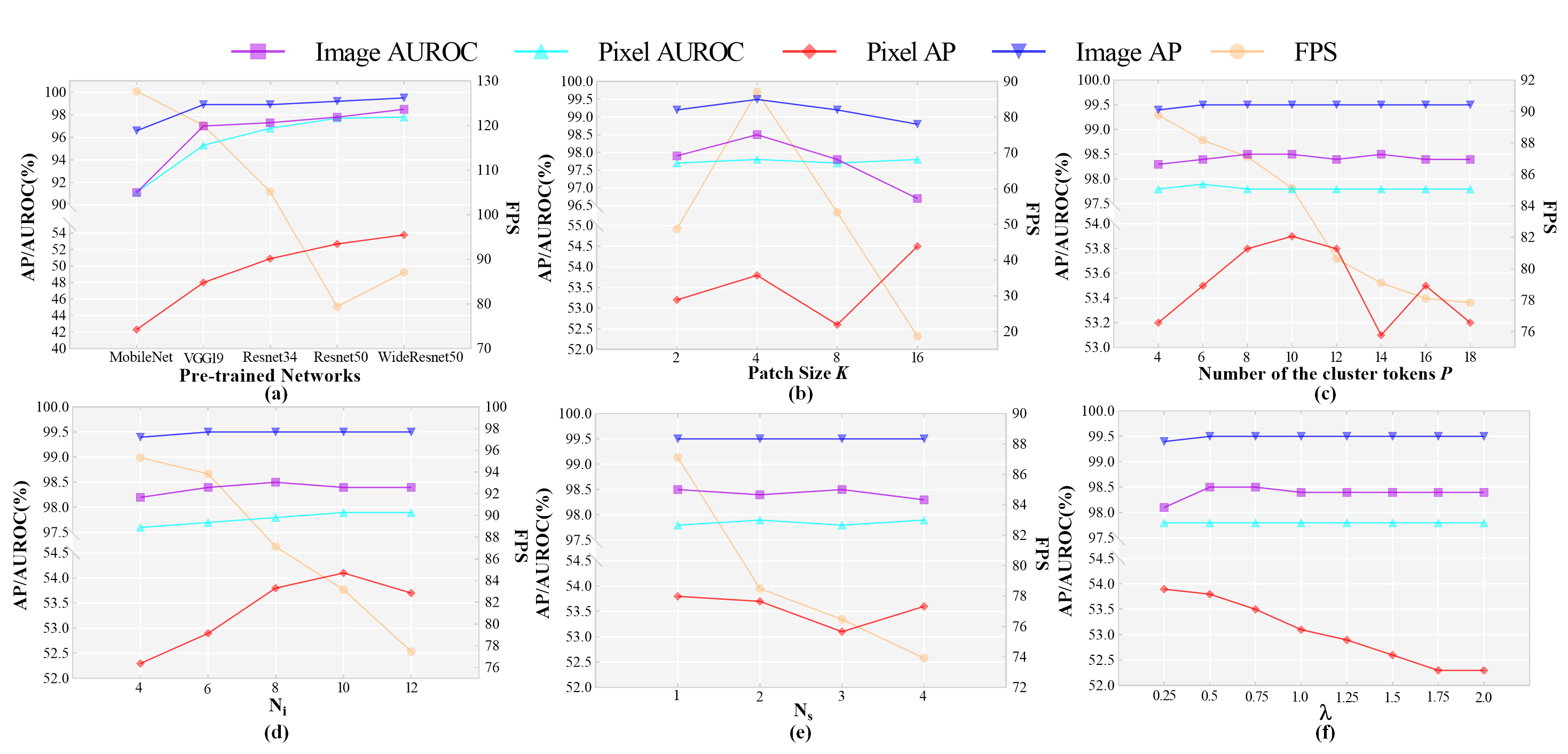}
    \caption{The ablation experiment results. (a)  Influence of pre-trained network. (b) Impact of patch size $K$. (c) Influence of the number of cluster tokens $P$. (d) Influence of the number of transformer block in inpainting network $N_i$. (e) Influence of the number of transformer block in semantic information aggregation network $N_s$. (f) Influence of the scaling factor $\lambda$.}
    \label{fig:ablation}
\end{figure*}
\subsubsection{Impact of Pre-trained Network}
\label{pre-trianednetwork}
Pre-trained networks are employed for extracting features from images. AMI-Net utilizes the extracted features as reconstruction targets; hence, the quality of the extracted features critically influences the model's performance. Consequently, we conduct comparative experiments on different variants of AMI-Net utilizing various pre-trained networks.\\
\indent The quantitative ablation experiment results are illustrated in Fig. \ref{fig:ablation}(a). As more intricate pre-trained networks are employed for feature extraction, the detection accuracy of the model exhibits an upward trend while its detection speed progressively diminishes. Notably, the model variant employing WideResnet50 outperforms its Resnet50 counterpart in both detection accuracy and speed. This distinction arises from the fact that WideResnet50 employs a broader and shallower convolutional architecture. As discussed earlier, industrial defect detection necessitates a favorable equilibrium between detection accuracy and speed. Consequently, we adopt WideResnet50 as the pre-trained network of choice to achieve this balance.

\subsubsection{Influence of Patch Size $K$}
The patch size $K$ influences the granularity of feature tokens, consequently affecting the quality of feature reconstruction. Therefore, a comprehensive analysis of the impact of $K$ on model performance and speed is provided below.\\
\indent Quantitative experimental results are illustrated in Fig. \ref{fig:ablation}(b). When the value of parameter $K$ is either excessively large or excessively small, it leads to a decline in the detection speed of the model. The former can be attributed to an excessive number of parameters in the linear projection, while the latter is due to an elongated token sequence, resulting in a significant increase in computational burden for the self-attention mechanism. Through our ablation experiments, it is observed that when $K$ is set to 4, the detection speed is optimized. Furthermore, this configuration yields the best results in terms of image/pixel-level AUROC and image-wise AP metrics, and the second-best performance in terms of pixel-wise AP metric. Therefore, considering a balanced trade-off between detection speed and accuracy, the patch size $K$ for the AMI-Net architecture is set to 4.

\subsubsection{Influence of the Number of Cluser Tokens $P$}
The number of cluster tokens $P$ significantly influences the clustering learning process, consequently impacting the model's performance. Hence, we conduct a comprehensive analysis of the effects of varying $P$ on the model's performance.\\
\indent The quantitative analysis results regarding the parameter $P$ are depicted in Fig. \ref{fig:ablation}(c). An increase in $P$ leads to a rise in computational demand for the network, subsequently resulting in a slowdown of the model's detection speed. On the other hand, when $P$ is too small, the model fails to capture a sufficient amount of normal information. Through our ablation experiments, we observe that models with $P$ set to 8 and 10 exhibit high detection accuracy. However, the model with $P$ set to 8 outperforms the variant with $P$ set to 10 significantly in terms of detection speed. Consequently, considering the equilibrium between detection speed and accuracy, we set the value of $P$ for the model to 8.
\subsubsection{Influence of the $N_i$}
The $N_i$ represents the number of transformer blocks within the inpainting network. The magnitude of $N_i$ directly influences the model's restoration capability, consequently impacting the model's detection performance. Hence, we conduct an elaborate analysis of the effects of varying $N_i$ on the model's performance.\\
\indent The quantitative results of ablation experiments regarding the impact of $N_i$ on model performance are presented in Fig. \ref{fig:ablation}(d). With an increase in the value of $N_i$, the detection speed of the model consistently decreases, while the detection performance demonstrates an ascending trend until reaching saturation. In our ablation experiments, model variants with $N_i$ set to 8 and 10 exhibited optimal outcomes in terms of image/pixel AUROC and image AP metrics. The model variant with $N_i$ equal to 10 slightly outperformed the $N_i$ equal to 8 variant in terms of pixel AP metric, while the $N_i$ equal to 8 variant significantly outperformed the $N_i$ equal to 10 variant in terms of detection speed. Consequently, considering the balance between detection speed and accuracy, we set the value of $N_i$ for the model to 8.

\subsubsection{Influence of the $N_s$}
The $N_s$ represents the number of transformer blocks within the semantic information aggregation network. The magnitude of $N_s$ influences the learning process of clustering, consequently impacting the detection accuracy of the model. Therefore, we conduct a comprehensive analysis of the effects of varying $N_s$ on the model's performance.\\
\indent Quantitative experimental results are presented in Fig. \ref{fig:ablation}(e). As $N_s$ increases, both the detection speed and accuracy of the model exhibit a declining trend. This phenomenon arises because during testing, when $N_s$ is excessively large, the semantic aggregation network aggregates defect-related information into cluster tokens, consequently causing the defect features to be imprecisely detected. Hence, we set the value of $N_s$ in AMI-Net to 1. 
\subsubsection{Influence of the $\lambda$}
The $\lambda$ represents the scaling factor for clustering clusters, which is utilized to control the size of clustering boundaries. It directly influences the detection of defect features, thereby impacting the model's performance. Consequently, an ablation experiment concerning the effects of varying $\lambda$ on model performance is conducted as follows.\\
\indent The quantitative experimental results are presented in Fig. \ref{fig:ablation}(f). The variation in parameter $\lambda$ does not affect the pixel AUROC metric, likely due to pixel AUROC favoring larger area defects, resulting in imprecise evaluations. With an increase in the value of $\lambda$, image AUROC displays a trend of initial increase followed by decrease, image AP metric shows a trend of initial increase followed by saturation, and pixel AP metric exhibits a decreasing trend. Considering all the metrics, we find that the model's overall performance is optimal when $\lambda$ is set to 0.5.

\begin{figure}[!t]
    \centering
    \includegraphics[width=88mm]{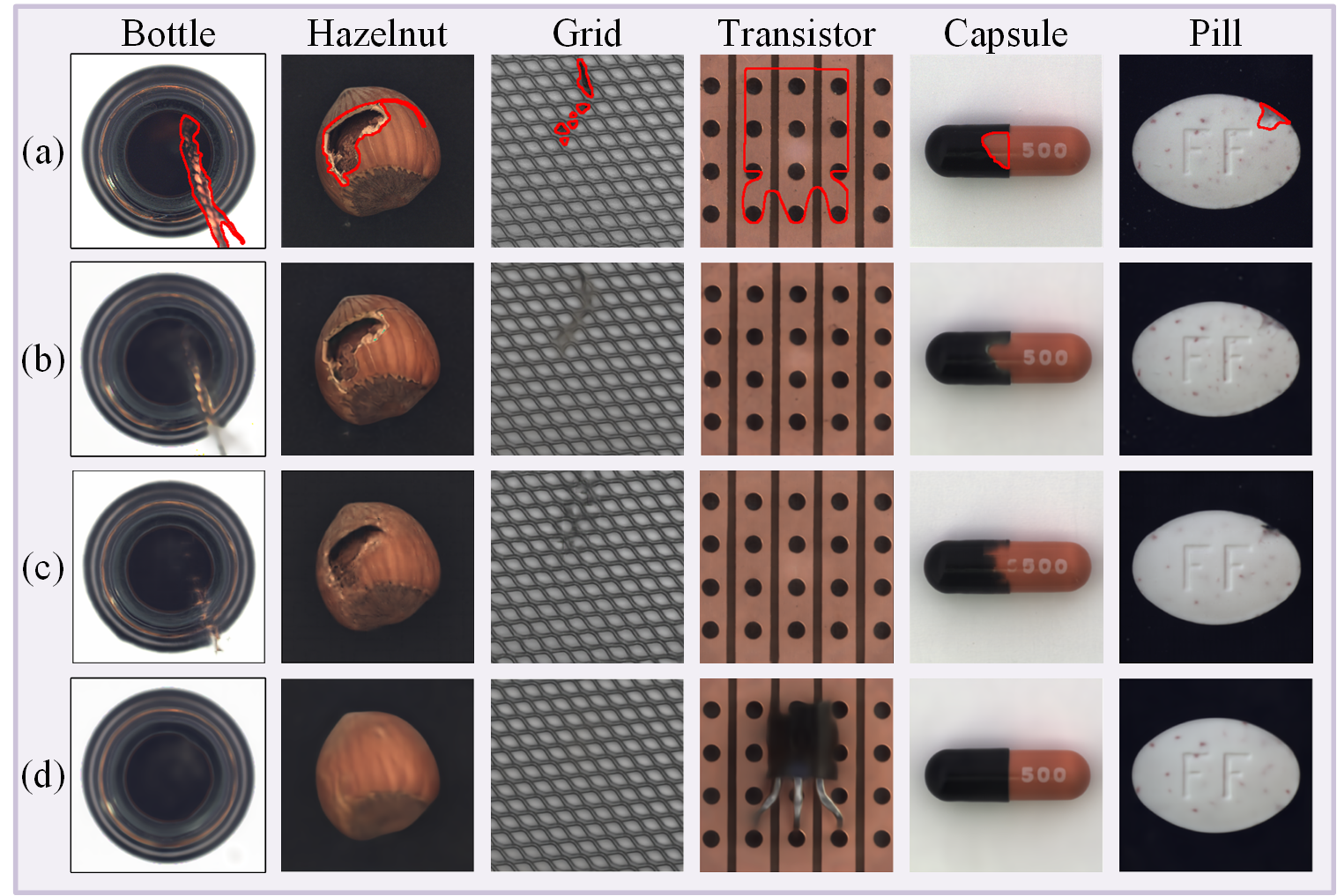}
    \caption{Reconstructed results of various methods. (a) Input defect image. (b) Reconstructed results with the vanilla autoencoder \cite{AE}. (c) Reconstructed results with existing mask-based method \cite{RIAD}. (d) Reconstructed results with the proposed AMI-Net. It is noteworthy that our model is learned for feature reconstruction and a separate
    decoder is employed to render images from features. This decoder is only used for visualization.}
    \label{fig:visrec}
\end{figure}
\subsection{Visualization Results}
As illustrated in Fig. \ref{fig:visrec}, we visualize the reconstruction results of different methods. The vanilla autoencoder perfectly reconstructs defects, given its lack of an explicit defect feature suppression mechanism. While existing mask-based methods can partially suppress the reconstruction of defects to some extent, the use of random masks during the testing process leads to the leakage of defect information, resulting in the reconstruction of a portion of defects. In contrast, our approach employs adaptive masks that entirely cover the defect regions, preventing the leakage of defect information and ensuring that defects are no longer reconstructed. The visualized reconstruction results vividly demonstrate the superiority of our method.
\begin{figure}[!t]
    \centering
    \includegraphics{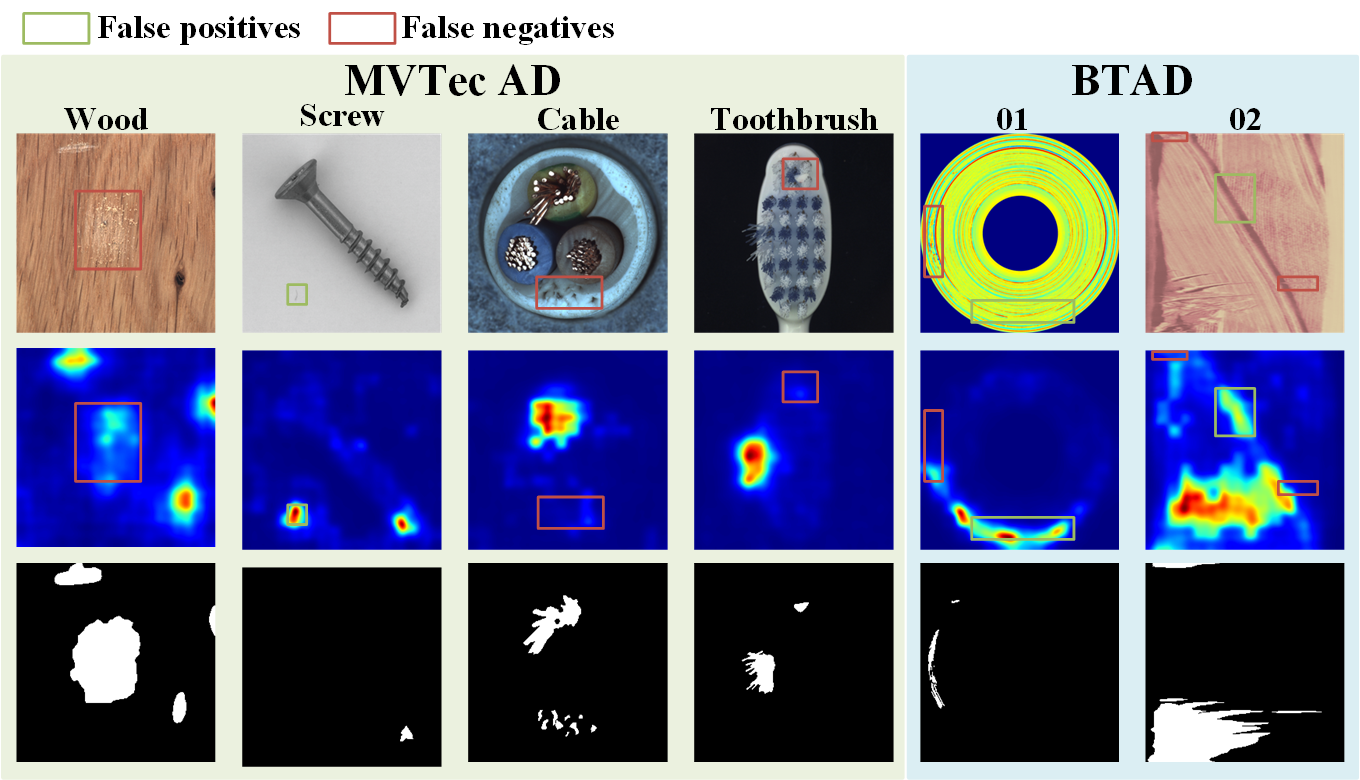}
    \caption{Some instances of detection failures by AMI-Net. Top Row: the defective sample. Mid Row: The detection result. Bottom Row: The corresponding label.}
    \label{fig:fasle-ex}
    \vspace{-1em}
\end{figure}
\subsection{Failure Case Analysis}
The extensive array of experiments presented above corroborates the efficacy and superiority of the proposed AMI-Net. However, it is important to acknowledge that AMI-Net does come with inherent limitations.\\
\indent Fig. \ref{fig:fasle-ex} illustrates a selection of instances where AMI-Net encounters detection failures. When defects resemble the texture of the background or manifest as subtle anomalies, AMI-Net struggles to differentiate them, leading to false negatives, as evident in cases such as Wood and Cable. Additionally, when minor background noise is present, AMI-Net can mistakenly detect it as defects, resulting in false positives, as observed in cases such as Screw and Class 02. These instances of misclassification stem from the model's insufficient capability to discern defects. Therefore, our next research endeavor aims to enhance the model's proficiency in defect discrimination.

\subsection{Discussion on Extending AMI-Net to Cases with Some Abnormal Training Samples.}
In real industrial settings, the quantity of normal samples significantly surpasses that of abnormal samples. Consequently, the proposed AMI-Net utilizes only normal samples for training. In fact, a few abnormal samples are often available in real-world industrial scenarios, the valuable knowledge of known defects should also be effectively exploited. Therefore, in this subsection, we discuss the extension of our AMI-Net to address this particular situation.\\
\indent AMI-Net is a reconstruction-based approach; thus, direct utilization of abnormal training samples is not feasible due to the absence of corresponding normal reconstruction targets. Drawing inspiration from BGAD \cite{BGAD}, we propose extracting defect regions from abnormal training samples, subjecting them to data augmentation, and subsequently randomly pasting them onto normal samples. This process yields more authentic artificial defect samples, with their reconstruction targets corresponding to the respective normal samples.\\
\indent Of course, the aforementioned strategy is merely a preliminary approach. In future research, we aim to refine this methodology continuously, enabling AMI-Net to more effectively extend to cases with some abnormal training samples.

\section{Conclusion}
This paper introduces an innovative AMI-Net designed for precise anomaly detection. Addressing the challenge of significant variations in industrial defect sizes, AMI-Net incorporates a strategy involving random positional and quantitative masks during training, enabling it to learn how to handle defects of various sizes. Subsequently, during testing, AMI-Net employs an adaptive mask generator to create specific masks that obscure defect regions while preserving normal areas. The inpainting network then utilizes available normal information to restore the masked regions. Extensive performance comparisons and ablation experiments validate the effectiveness and superiority of AMI-Net. However, our research has identified that AMI-Net exhibits reduced capabilities in detecting subtle defects and those with low contrast. In our future investigations, we will concentrate on addressing this limitation.

\bibliographystyle{IEEEtran}
\bibliography{ref}

\begin{IEEEbiography}[{\includegraphics[width=1in,height=1.25in,clip,keepaspectratio]{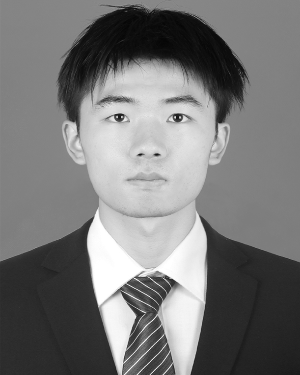}}]{Wei Luo}(Student Member, IEEE) received a B.S. degree from the School of Mechanical Science and Engineering, Huazhong University of Science and Technology, Wuhan, China, in 2023. He is pursuing a Ph.D. degree with the Department of Precision Instrument, Tsinghua University.\\
\indent His research interests include deep learning, anomaly detection and machine vision.
\end{IEEEbiography}
\vspace{1\baselineskip}

\begin{IEEEbiography}[{\includegraphics[width=1.25in,height=1.25in,clip,keepaspectratio]{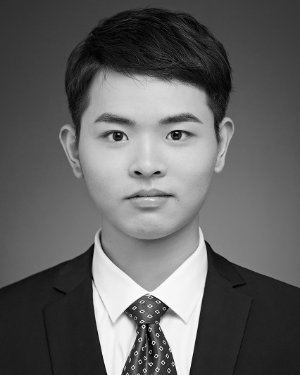}}]{Haiming Yao}(Student Member, IEEE)
received a B.S. degree from the School of Mechanical Science and Engineering, Huazhong University of Science and Technology, Wuhan, China, in 2022. He is pursuing a Ph.D. degree with the Department of Precision Instrument, Tsinghua University.\\
\indent His research interests include deep learning, edge intelligence and machine vision.
\end{IEEEbiography}
\vspace{1\baselineskip}

\begin{IEEEbiography}[{\includegraphics[width=1in,height=1.25in,clip,keepaspectratio]{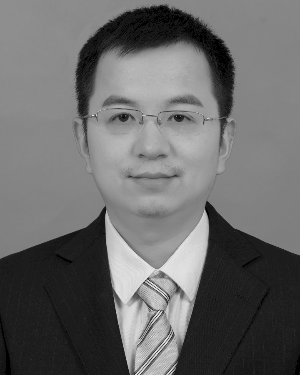}}]{Wenyong Yu}(Senior Member, IEEE)
received an M.S. degree and a Ph.D. degree from Huazhong University of Science and Technology, Wuhan, China, in 1999 and 2004, respectively. He is currently an Associate Professor with the School of Mechanical Science and Engineering, Huazhong University of Science and Technology.\\
\indent His research interests include machine vision, intelligent control, and image processing.
\end{IEEEbiography}
\vspace{1\baselineskip}

\begin{IEEEbiography}[{\includegraphics[width=1in,height=1.25in,clip,keepaspectratio]{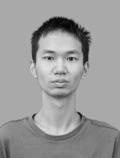}}]{Zhengyong Li}
received a B.S. degree from the School of Mechanical Science and Engineering, Huazhong University of Science and Technology, Wuhan, China,in 2023. He is going to pursue a M.S. degree in Mechanical Engineering, Huazhong University of Science and Technology.\\
\indent His research include machine vision, deep learning and defect detection.
\end{IEEEbiography}
\vspace{1\baselineskip}
\end{document}